\documentclass[11pt]{article}

\usepackage[preprint]{acl}

\usepackage{times}
\usepackage{latexsym}

\usepackage[T1]{fontenc}

\usepackage[utf8]{inputenc}

\usepackage{microtype}

\usepackage{inconsolata}

\usepackage{graphicx}

\usepackage{amsmath}      
\usepackage{amssymb}      
\usepackage{amsfonts}     
\usepackage{bm}           
\usepackage{algorithm}    
\usepackage{algorithmic}  
\usepackage{float}        

\usepackage{amsthm}
\newtheorem{proposition}{Proposition}

 \usepackage{enumitem}

\usepackage{tabularx}
\usepackage[table]{xcolor}
\usepackage{multirow}
\usepackage{hyperref}

\usepackage{makecell} 
\newcolumntype{Y}{>{\centering\arraybackslash}X}

\usepackage[utf8]{inputenc}
\usepackage[T1]{fontenc}
\usepackage{xcolor}
\usepackage[skins,breakable]{tcolorbox}
\definecolor{dustyblue}{HTML}{728FA5}
\newtcolorbox{genbox}[1][]{
  enhanced,
  breakable,
  colback=dustyblue!10, 
  colframe=dustyblue!90!black, 
  boxrule=0.5mm,
  arc=3mm,
  left=3mm, right=3mm, top=2mm, bottom=2mm,
  fonttitle=\bfseries\sffamily,
  coltitle=white,
  title={#1},
  shadow={2mm}{-1mm}{0mm}{black!20}
}
\newcommand{\loopmark}{\textcolor{dustyblue!80!black}{\textit{[collapsed loop]}}}

\title{The More Popular, The Harder to Forget:\\ Adaptive Popularity for LLM Unlearning}

\author{
  \textbf{Anna Borisiuk\textsuperscript{1,4}}\thanks{\ \texttt{a8or1suk@gmail.com}},
  \textbf{Andrey Savchenko\textsuperscript{2,4}},
  \textbf{Alexander Panchenko\textsuperscript{3,1}}, 
  \textbf{Elena Tutubalina\textsuperscript{1,4}}\\
  \textsuperscript{1}AIRI, 
  \textsuperscript{2}Sber AI Lab, 
  \textsuperscript{3}Skoltech, 
  \textsuperscript{4}ISP RAS Research Center for Trusted Artificial Intelligence \\ 
  }
\begin{document}
\maketitle

\begin{abstract}


Popular facts are memorised more deeply during pretraining and resist removal longer than rare ones, yet existing LLM unlearning methods apply uniform gradient pressure regardless of training-data frequency. We propose the \textbf{AdaPop} (\textbf{Ada}ptive \textbf{Pop}ularity) method, which combines local token confidence with a per-fact popularity-dependent exponent derived from an external proxy (e.g., Wikidata sitelinks, LLM-as-Judge), and automates the forget-retain balance via a dual-ascent controller that adjusts the retain penalty each epoch. Across three model families and two benchmarks, AdaPop leaks \(\sim 5\times\) less forgotten content than competing methods under paraphrased queries and \(\sim 1.6\times\) less under adversarial reformulations. We support our analysis with internal metrics: under our method, forget-set hidden states move further from the pre-unlearning model's states than under other methods, while retain-set representations remain close.

\end{abstract}

\section{Introduction}
\label{sec:intro}
Machine unlearning (MU)~\cite{cao2015towards, bourtoule2021machine} removes targeted knowledge from a trained model without full retraining, addressing privacy, safety, and compliance risks that arise when LLMs memorise pretraining data~\cite{carlini2021extracting, tirumala2022memorization}. The forget set holds the (query, answer) pairs to remove; a held-out retain set measures preserved capability.

Existing gradient-based and preference-learning objectives~\cite{jang2022knowledge, liu2022continual, zhang2024negative, wang2025rethinking} treat every forget example as equally hard to erase. This assumption is empirically false: facts seen more often during pretraining are more deeply encoded and harder to remove~\cite{kandpal2023large, merullo2025linear}. Applying uniform gradient magnitude across the forget set therefore produces a systematic failure mode, the \emph{popularity gap}: rare facts are over-erased, damaging retain quality, while popular facts are under-erased and remain recoverable under paraphrase or adversarial prompts. Recent work documents this gap across methods and scales~\cite{krishnannot, anna2026anatomy} but proposes no training-time remedy. \citet{yu2026falw} do reweight at training time, though in long-tailed image classification and from the model's own predictive probability (§\ref{subsec:popularity_effects}). Methods that calibrate the unlearning signal from model-intrinsic quantities (e.g., per-token confidence in~\citealt{wang2025rethinking}) cannot resolve the gap, because confidence reflects current output behaviour rather than parametric encoding depth.

\begin{figure*}[t]
    \centering
    \includegraphics[width=0.99\textwidth]{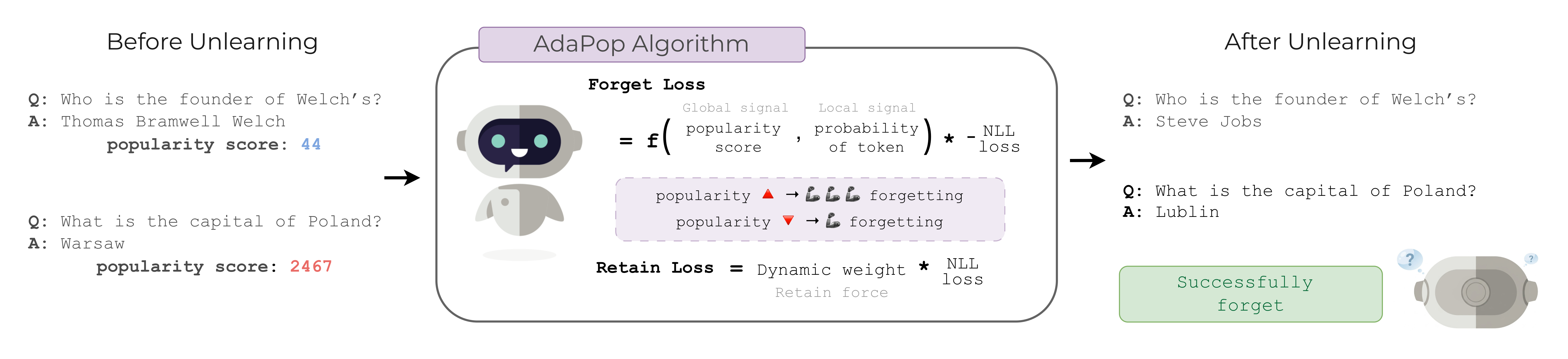}
    \caption{Overview of AdaPop. The popularity score reweights per-fact token contributions to the forget loss; the dual-ascent controller adjusts the retain penalty each epoch to maintain retain quality (Algorithm~\ref{alg:adapop}).}
    \label{fig:adapop_workflow}
\end{figure*}

We introduce \textbf{AdaPop} (\textbf{Ada}ptive \textbf{Pop}ularity), the first training-time unlearning method to condition per-fact gradient updates on an \emph{external} popularity signal (Figure~\ref{fig:adapop_workflow}). AdaPop converts a per-fact popularity score into a power-law exponent that sharpens or softens the ascent signal on each token, and pairs this reweighting with a dual-ascent controller that adapts the retain penalty online. We evaluate across Llama~\cite{grattafiori2024llama}, Qwen~\cite{qwen2025}, and Gemma~\cite{team2024gemma} on two real-world benchmarks~\cite{anna2026anatomy, jin2024rwku}, using MMLU~\cite{hendrycksmeasuring} and HellaSwag~\cite{zellers2019hellaswag} to track general capability retention. Our contributions are:
\begin{itemize}
    \item \textbf{Popularity-aware gradient weighting.} AdaPop maps each fact's external popularity (e.g., Wikidata-based proxy scores, LLM-as-a-judge estimates) to a per-fact power-law exponent that reweights its token-level ascent signal. This targets the popularity gap directly rather than through model confidence.
    \item \textbf{Automated forget-retain balance.} An epoch-level dual-ascent controller treats the retain penalty as a Lagrange multiplier, adjusting it in response to observed retain-loss drift.
    This eliminates per-dataset hyperparameter search and prevents retain collapse under increasing forget pressure.
    \item \textbf{Empirical evaluation.} Among methods that do not collapse, AdaPop attains the lowest adversarial-query scores on all three models and the best paraphrase scores on two of three. Token rankings and hidden states show that the answer is disrupted inside the model rather than only hidden at the output, while general capabilities stay within $0.05$ MMLU of the pre-unlearning checkpoint.
\end{itemize}

We've release our code in a GitHub repository\footnote{\url{https://github.com/Anya-wUw/open-unlearning}}.

\section{Related Work}
\label{sec:related}
\subsection{Machine Unlearning Methods}
\label{subsec:gradient_unlearning}

Gradient-based unlearning methods fall into two families. \emph{Ascent-based} methods maximise the negative log-likelihood on the forget set: Gradient Ascent (GA;~\citealt{jang2022knowledge}) does so without constraint and often collapses linguistic coherence~\cite{yao2024large, zhang2024negative}; Gradient Difference (GD;~\citealt{liu2022continual}) adds a retain loss but treats every forget token uniformly~\cite{bu2025unlearning}; Weighted Gradient Ascent (WGA;~\citealt{wang2025rethinking}) weights each token's loss by the model's confidence on that token, which makes every token contribute equally to the gradient. \emph{Preference-based} methods such as Negative Preference Optimization (NPO;~\citealt{zhang2024negative}) push the output distribution away from the memorised response. Ascent-based methods forget aggressively but need retain regularisation; preference-based methods are gentler but converge more slowly on deeply memorised facts, a trade-off we revisit empirically (§\ref{sec:results}). None of the four draws on a signal from outside the model: the unlearning pressure follows entirely from the model's current output distribution, which does not reflect parametric encoding depth, a limitation AdaPop directly addresses (§\ref{sec:AdaPop}, Table~\ref{tab:robustness}). \citet{lucki2025adversarial} observe that even inference-time interventions targeting outputs leave parametric knowledge intact.

\subsection{Memorisation, Frequency, and Unlearning}
\label{subsec:memorization}

LLM memorisation scales with training-corpus frequency~\cite{carlini2023quantifying, tirumala2022memorization}: \citet{kandpal2023large} show LLMs recall long-tail facts unreliably, and \citet{merullo2025linear} demonstrate that frequent facts are encoded more linearly, providing a mechanistic basis for why popular facts resist gradient-based removal. This asymmetry works against unlearning: verbatim memorisation impedes removal in both vision~\cite{NEURIPS2024_16e18fa3} and LLM settings~\cite{barbulescu2024each}, and \citet{baluta2024unlearning} show that differently exposed data points unlearn differently. Log-probabilities do not reflect memorisation depth~\cite{wang2025rethinking, hartmann2024undesirable}: a model may assign high confidence to a surface pattern even when the underlying fact is shallowly encoded. External proxies such as sitelink counts give a fact-level signal independent of the model's output distribution.

\subsection{Popularity Effects on Unlearning}
\label{subsec:popularity_effects}

\citet{krishnannot} show through controlled pretraining experiments that high-frequency facts are either not unlearned or only superficially forgotten, remaining extractable via adversarial queries. \citet{anna2026anatomy} establish the same failure mode on real-world facts in DUET, showing that Wikipedia-based fact salience predicts unlearning difficulty across existing methods. Both works characterise the popularity gap and provide the benchmarks we build on, but neither proposes a training method to close it. We address this gap by introducing a new adaptive popularity method (§\ref{sec:AdaPop}).

\paragraph{Local versus external calibration.}
Reweighting the forget signal is not itself new; what separates methods is the quantity the weight is computed from. WGA~\cite{wang2025rethinking} and FaLW~\cite{yu2026falw} both derive it from the model's own predictive probability, a \emph{local} signal that shifts as training proceeds. AdaPop derives its per-fact exponent from an \emph{external} corpus-frequency proxy fixed before training and independent of the model's outputs. The distinction is empirical rather than terminological: a local weight reports what the model currently emits, so it saturates once the surface answer is suppressed, which is precisely where WGA's forget $\Delta$Rank turns negative while paraphrase recall survives (§\ref{sec:internal}). FaLW further requires a reference distribution over unseen same-class data; in fact-level unlearning each fact is effectively its own class, so no such distribution exists.

\section{Adaptive Popularity Method}
\label{sec:AdaPop}
We frame targeted unlearning as a constrained optimisation: increase the forget-set loss while bounding retain drift. A single retain coefficient $\alpha$ shared across all facts (GD) or a confidence-based per-token weight (WGA) cannot serve both popularity regimes: at a single gradient magnitude, popular facts resist erasure while rare facts incur collateral damage~\cite{hartmann2024undesirable, kandpal2023large}. AdaPop instead combines (i) a per-fact power-law exponent computed from an external popularity proxy, which reweights per-token ascent signals, and (ii) a dual-ascent retain controller that updates the retain penalty each epoch.

\subsection{Problem Formulation}

Let $\mathcal{D}$ be the pretraining dataset on which the model $q_\theta$ has been trained. We identify two disjoint subsets $\mathcal{D}_F, \mathcal{D}_R \subset \mathcal{D}$ ($\mathcal{D}_F \cap \mathcal{D}_R = \emptyset$, $\mathcal{D}_F \cup \mathcal{D}_R \subsetneq \mathcal{D}$): the \emph{forget} set $\mathcal{D}_F=\{(x_i,y_i,s_i)\}_{i=1}^{N_F}$, where each entry couples a question-answer pair $(x_i, y_i)$ with a popularity score $s_i$ from a proxy such as the Wikidata sitelink score (§\ref{sec:experiments}). The \emph{retain} set is $\mathcal{D}_R=\{(x_j,y_j)\}_{j=1}^{N_R}$, a held-out sample of knowledge the model must preserve. We consider an autoregressive model $q_\theta$ defined by the conditional probability $q_\theta(y_t \mid x, y_{<t})$ of a token $y_t$ given its prefix. For a token position $t$ in sequence $i$, the negative log-likelihood (NLL) is
\begin{equation}
    \ell_{i,t}(\theta) = -\log q_\theta(y_{i,t} \mid x_i, y_{i,<t}),
\end{equation}
and the model's confidence (a \emph{local} signal from the model's current output distribution) is
\begin{equation}
    p_{i,t}(\theta) = \exp(-\ell_{i,t}(\theta)).
\end{equation}

$\Omega_{\mathcal{F}}, \Omega_{\mathcal{R}}$ index the valid (non-padding)
answer tokens of $\mathcal{D}_F, \mathcal{D}_R$. The retain loss is the standard supervised NLL:
\begin{equation}
\label{eq:retain_loss}
L_r(\theta) = \frac{1}{|\Omega_{\mathcal{R}}|} \sum_{(j,t) \in \Omega_{\mathcal{R}}} \ell_{j,t}(\theta).
\end{equation}
The per-token weight $w_{i,t}$ (§\ref{subsec:popularity-obj}) combines local confidence with the popularity score $s_i$, an \emph{external} signal from the pretraining corpus. AdaPop maximises forgetting on $\mathcal{D}_F$ subject to retain drift $\delta_k \le \varepsilon$, enforced by a dual-ascent controller (§\ref{sec:dual_ascent}) rather than the per-dataset grid search a fixed $\alpha$ requires. The design rationale is in Appendix~\ref{app:design_motivation}.

\subsection{Popularity-Aware Forget Objective}
\label{subsec:popularity-obj}

Each sample $i$ in the forget set is assigned an exponent $\beta_i$ based on its popularity score $s_i$:
\begin{equation}
    \beta_i = a \cdot s_i^{-b}.
\end{equation}
The power-law mapping mirrors the empirical observation that parametric memorisation scales as a power law with corpus frequency~\cite{kandpal2023large}, and compresses the proxy's wide dynamic range into a bounded exponent. Because $\beta_i$ is continuous and strictly decreasing in $s_i$, a more popular fact always receives more \emph{pressure-sustaining} weighting ($\beta$ smaller, amplifying gradient mass as tokens are erased). We fix $a$ and $b$ from two anchor scores taken at the rare and popular ends of the DUET score distribution ($s_r \approx 100$ and $s_p \approx 3{,}000$; Figure~\ref{fig:benchmarks_pop_diverse}), requiring the rare anchor to receive $\beta = 1.5$, the \emph{self-limiting} regime in which the gradient on a token attenuates as it is forgotten so erasure halts on its own, and the popular anchor $\beta = 0.1$, well inside the pressure-sustaining regime. These two conditions determine $(a, b)$ in closed form, and only the ratio $s_p/s_r$ enters $b$, so the anchors need order-of-magnitude placement rather than precise distributional statistics (Appendix~\ref{app:derivation}). Exponents are clipped to a bounded interval so that outlier scores cannot produce extreme gradients.

The per-token unlearning weight is
\begin{equation}
\label{eq:weight}
w_{i,t} = \mathrm{sg}\!\left[p_{i,t}(\theta)^{\beta_i}\right],
\end{equation}
and the forget loss is
\begin{equation}
\label{eq:forget_loss}
L_f(\theta) = -\frac{1}{|\Omega_{\mathcal{F}}|} \sum_{(i,t) \in \Omega_{\mathcal{F}}} w_{i,t}\, \ell_{i,t}(\theta).
\end{equation}
Here $\mathrm{sg}[\cdot]$ is a stop-gradient. The weight $w_{i,t}$ is applied only during training, only over answer tokens; question tokens are masked. Because $w_{i,t}$ carries no gradient, the backward pass runs through $\ell_{i,t}$ alone, and $\nabla_\theta \ell_{i,t} = -\nabla_\theta p_{i,t} / p_{i,t}$ contributes a factor $1/p_{i,t}$. The per-token gradient magnitude is therefore $p_{i,t}^{\beta_i} \cdot p_{i,t}^{-1} = p_{i,t}^{\beta_i-1}$, one power lower than the weight itself. This makes $\beta{=}1$ the crossover: above it the gradient vanishes as $p_{i,t}\to 0$, below it the gradient grows, and at $\beta{=}1$ every token contributes uniformly, recovering WGA. Appendix~\ref{app:derivation:regime} states and proves this as Proposition~\ref{prop:regimes}. The combined training loss is
\begin{equation}
\label{eq:total_loss}
L(\theta; \alpha) = L_f(\theta) + \alpha L_r(\theta),
\end{equation}
optimised by gradient descent on $\theta$, which performs ascent on the weighted forget NLL and descent on the retain NLL in a single update. The coefficient $\alpha$ is set by the dual-ascent controller below.

\subsection{Dual-Ascent and Auto-Balancing}\label{sec:dual_ascent}

AdaPop treats $\alpha = \alpha_0 + \lambda$ as a dual variable enforcing $\delta_k \le \varepsilon$, where $\delta_k$ is the one-sided relative drift of the retain loss at epoch $k$:
\begin{equation}
\delta_k = \max\!\left( 0,\; \frac{R_k - R_{\mathrm{ref}}}{\max(R_{\mathrm{ref}}, \xi)} \right).
\end{equation}
$R_k = L_r(\theta_k)$ is the retain loss at the end of epoch $k$, $R_{\mathrm{ref}} = R_1$ is the reference value from epoch 1, and $\xi > 0$ is a numerical-stability constant. Only positive drift triggers the constraint.

\paragraph{Epoch-level dual update.}
The dual variable is updated by projected gradient ascent once per epoch (not per step), suppressing oscillations from noisy per-batch retain losses:
\begin{align}
    \lambda_{k+1} &= \Pi_{[0, \lambda_{\max}]}\!\left( \lambda_k + \eta_\lambda (\delta_k - \varepsilon) \right) \\
    \alpha_{k+1} &= \alpha_0 + \lambda_{k+1}.
\end{align}
When $\delta_k > \varepsilon$, $\lambda$ grows and $\alpha$ tightens the retain constraint; when $\delta_k \le \varepsilon$, $\lambda$ decays and the forget signal acts more aggressively. Because $\alpha$ scales the retain penalty against the popularity-weighted forget loss in Eq.~\eqref{eq:total_loss}, the feedback governs the trade-off between popularity-calibrated ascent and retain preservation, not between uniform ascent and retain preservation. Appendix~\ref{app:ablation} shows that removing either the controller or the popularity exponent leaves a distinct failure mode unaddressed.


\begin{algorithm}[ht!]
\caption{AdaPop}
\label{alg:adapop}
\begin{algorithmic}[1]
\REQUIRE Forget set $\mathcal{D}_F=\{(x_i,y_i,s_i)\}$, retain set $\mathcal{D}_R$, model $q_\theta$, initial retain weight $\alpha_0$, dual step size $\eta_\lambda$, tolerance $\varepsilon$, dual bound $\lambda_{\max}$, anchors $(s_r, s_p)$, clip range $[\beta_{\min}, \beta_{\max}]$
\ENSURE Updated parameters $\theta$

\STATE Derive $a, b$ from anchor constraints $\beta(s_r), \beta(s_p)$ (Appendix~\ref{app:derivation})
\STATE Compute popularity exponents: $\beta_i \leftarrow \mathrm{clip}\big(a \cdot s_i^{-b},\, \beta_{\min},\, \beta_{\max}\big)$ for all $(x_i, y_i, s_i) \in \mathcal{D}_F$

\STATE Initialise $\lambda \leftarrow 0$, $\alpha \leftarrow \alpha_0$, $R_{\mathrm{ref}}$ unset

\FOR{epoch $k = 1$ to $K$}

    \FOR{batch $B$ sampled from $\mathcal{D}_F \cup \mathcal{D}_R$}
        \STATE Let $B_F = B \cap \mathcal{D}_F$, $B_R = B \cap \mathcal{D}_R$, with answer-token indices $\Omega_{B_F}, \Omega_{B_R}$
        \FOR{forget sample $i \in B_F$ and answer token $t$}
            \STATE $p_{i,t} \leftarrow \exp(-\ell_{i,t}(\theta))$
            \STATE $w_{i,t} \leftarrow \mathrm{sg}\!\left[p_{i,t}^{\beta_i}\right]$
        \ENDFOR

        \STATE $L_f \leftarrow -\dfrac{1}{|\Omega_{B_F}|}\sum\limits_{(i,t)\in \Omega_{B_F}} w_{i,t}\,\ell_{i,t}(\theta)$ \hfill$\triangleright$ Eq.~\eqref{eq:forget_loss}

        \STATE $L_r \leftarrow \dfrac{1}{|\Omega_{B_R}|}\sum\limits_{(j,t)\in \Omega_{B_R}} \ell_{j,t}(\theta)$ \hfill$\triangleright$ Eq.~\eqref{eq:retain_loss}

        \STATE $L \leftarrow L_f + \alpha L_r$ \hfill$\triangleright$ Eq.~\eqref{eq:total_loss}

        \STATE $\theta \leftarrow \theta - \eta\,\nabla_\theta L$ \hfill$\triangleright$ ascent on forget, descent on retain

    \ENDFOR

    \STATE $R_k \leftarrow L_r(\theta)$ over $\mathcal{D}_R$
    \IF{$k = 1$}
        \STATE $R_{\mathrm{ref}} \leftarrow R_k$
    \ENDIF

    \STATE $\delta_k \leftarrow \max\!\left(0,\;\dfrac{R_k - R_{\mathrm{ref}}}{\max(R_{\mathrm{ref}},\, \xi)}\right)$
    \STATE $\lambda \leftarrow \Pi_{[0,\lambda_{\max}]}\!\big(\lambda + \eta_\lambda(\delta_k - \varepsilon)\big)$
    \STATE $\alpha \leftarrow \alpha_0 + \lambda$

\ENDFOR

\RETURN $\theta$
\end{algorithmic}
\end{algorithm}

\section{Experimental Setup}
\label{sec:experiments}

\subsection{Baselines}
We evaluate \textbf{AdaPop} against four established unlearning baselines: GA~\cite{jang2022knowledge}, GD~\cite{liu2022continual}, NPO~\cite{zhang2024negative}, and WGA~\cite{wang2025rethinking}. Results for 11 additional methods (UNDIAL~\cite{dong2025undial}, RMU~\cite{li2024wmdp}, PDU~\cite{entesariconstrained}, NPO-SAM~\cite{fan2025towards}, SimNPO~\cite{fan2026simplicity}, SatImp~\cite{yangexploring}, Adaptive RMU~\cite{dang2025effects}, AltPO~\cite{mekala2025alternate}, FLAT~\cite{wang2025llm}, TPO~\cite{zhou2026not}, and CE-U~\cite{yang2025u}) are in Appendix~\ref{app:additional_baselines}. Main results use $\text{lr}=10^{-4}$, the stable operating point where most methods reach their best forgetting scores; a full sweep over five learning rates is in Appendix~\ref{app:lr_sensitivity} and per-epoch dynamics in Appendix~\ref{app:epoch_dynamics}. ROUGE-L is reported as mean\,$\pm$\,std over three random seeds. AdaPop hyperparameters are in Appendix~\ref{app:hyperparameters}; sensitivity to $\pm$20\% perturbations of $a$ and $b$ is in Appendix~\ref{app:coef_sensitivity}.

\subsection{Models}
We use Llama-3.1-8B-Instruct~\cite{grattafiori2024llama}, Gemma-7B-it~\cite{team2024gemma}, and Qwen2.5-7B-Instruct~\cite{qwen2025}, representing the 7--8B scale standard in unlearning research~\cite{zhang2024negative, yao2024large, jin2024rwku}. All experiments use LoRA fine-tuning~\cite{hu2021lora} ($r{=}32$, $\alpha_{\mathrm{LoRA}}{=}64$)~\citep{openunlearning2025}; full fine-tuning is excluded following \citet{anna2026anatomy}, who show it either fails to converge on popular facts or collapses into catastrophic forgetting. Per-model breakdowns are in Appendix~\ref{app:model_specific}.

\subsection{Benchmarks}
\begin{figure}[t]
    \centering
    \includegraphics[width=0.9\linewidth]{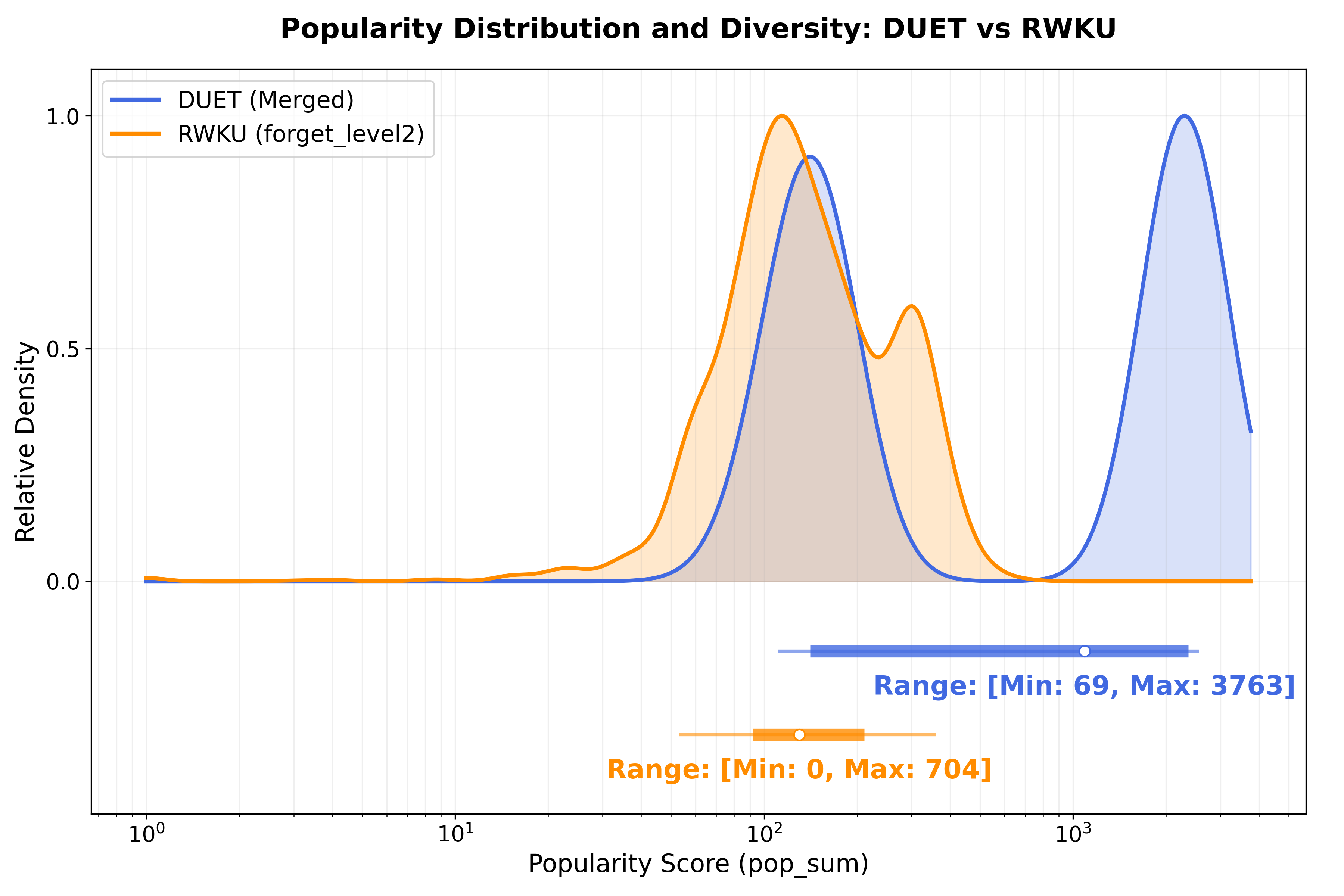}
    \caption{Wikidata popularity-score distributions for the forget sets of DUET and RWKU. DUET spans nearly two orders of magnitude ($69$--$3{,}763$; median $1{,}090$), providing a direct test of popularity-sensitive methods; RWKU is narrower and skewed lower (median $130$).}
    \label{fig:benchmarks_pop_diverse}
\end{figure}

We evaluate on two real-world knowledge datasets with different popularity distributions (Figure~\ref{fig:benchmarks_pop_diverse}):

    \textbf{DUET}~\cite{anna2026anatomy}: A city-centric factual QA benchmark with explicit popularity diversity. It covers a broad spectrum of Wikidata scores (range: 69--3{,}763; median: 1{,}090), enabling direct observation of the popularity gap. We use the combined rare-and-popular forget subset with a 500-sample fast retain set.
    
    \textbf{RWKU}~\cite{jin2024rwku}: A precision-focused benchmark testing targeted fact erasure while preserving semantically adjacent knowledge. Its popularity distribution is narrower and skewed lower (range: 0--704; median: 130), suitable for evaluating erasure under lower popularity variance.

We exclude synthetic benchmarks like TOFU~\cite{maini2024tofu}, where each persona has only 20 QA samples (score range $0$--$20$), too narrow to observe popularity effects. DUET provides this score natively; for RWKU we compute it from Wikidata sitelink counts following~\citet{anna2026anatomy}, which also tests stability across dataset-native and externally computed proxies. A usable proxy need only be continuous and span roughly an order of magnitude between rare and popular facts (Appendix~\ref{app:derivation}): by Proposition~\ref{prop:ratio} only the \emph{ratio} of the anchors enters $b$, so approximate placement suffices (§\ref{sec:proxy_ablation}).


\subsection{Evaluation Metrics}
We evaluate along two dimensions: output-level behaviour and internal representation change, following established protocols~\cite{xue2026towards, shimuse}. The second is essential because a model may suppress a correct answer at the surface while keeping the underlying fact retrievable under paraphrase or adversarial prompting, a failure mode we observe with WGA (\S\ref{sec:internal}). All internal metrics compare the unlearned model against the same checkpoint before unlearning; formal definitions are in Appendix~\ref{app:metric_formulas}.

\paragraph{Output-level metrics.}
Lower values on the forget set and higher values on the retain set indicate effective unlearning.
\begin{itemize}[leftmargin=1.2em,itemsep=1pt,topsep=2pt]
    \item \textbf{ROUGE-L Recall}~\cite{lin2004rouge}: token overlap between the generated output and the gold answer.
    \item \textbf{Cosine Similarity}~\cite{reimers2019sentence}: embedding-space proximity between generated and gold answer representations. A model that outputs a near-synonym scores low on ROUGE-L but high on cosine similarity, a signal of incomplete forgetting.
\end{itemize}

\paragraph{Internal metrics.}
These diagnose whether forgetting is genuine or merely surface suppression.
\begin{itemize}[leftmargin=1.2em,itemsep=1pt,topsep=2pt]
    \item \textbf{$\Delta$LP}: shift in the log-probability of the gold answer, summed over its tokens; large negative values on the forget set indicate the correct answer has become substantially less likely, while a retain-set value near zero ($\approx 0$) indicates the retained answers were left alone.
    \item \textbf{$\Delta$Rank}: change in the mean rank of the gold token in the predicted distribution; large positive values on the forget set indicate the correct answer has been demoted, a stronger erasure signal than $\Delta$LP alone. On the retain set the value should be near zero ($\approx 0$).
    \item \textbf{Hid.Cos}: cosine similarity between the answer-token hidden states of the last four layers, before and after unlearning; low values on the forget set signal representational disruption beyond output suppression.
    \item \textbf{KL}: KL divergence between the output distributions before and after unlearning; high values on the forget set indicate parametric forgetting.
\end{itemize}
Persistently high Hid.Cos, or a $\Delta$Rank on the forget set that is near zero or negative, flags surface suppression.

\section{Results}
\label{sec:results}

We structure the evaluation in five parts: output-level forgetting and retention (§\ref{sec:main_results}), robustness to paraphrase and adversarial queries (§\ref{sec:robustness}), internal representation analysis (§\ref{sec:internal}), general capability preservation (§\ref{sec:lm_eval}), and the popularity proxy ablation (§\ref{sec:proxy_ablation}).

\subsection{Main Results}
\label{sec:main_results}

Tables~\ref{tab:rouge-llama}--\ref{tab:rouge-gemma} report ROUGE-L Recall and Cosine Similarity. Across all three architectures and both benchmarks, \textbf{AdaPop} achieves, among stable (non-collapsed) methods, the lowest forget Cosine Similarity in all six (model $\times$ benchmark) cells and the lowest forget ROUGE-L in 5 of 6, with mean retain ROUGE-L $\geq 0.855$ on all model-benchmark combinations. The single exception is Llama/DUET, where WGA's forget ROUGE-L is marginally lower; on that cell AdaPop's deeper representational impact appears instead in Cosine Similarity and $\Delta$Rank (Table~\ref{tab:internal_agg}).

\begin{table}[ht]
\centering
\footnotesize
\setlength{\tabcolsep}{2.5pt}
\begin{tabularx}{\linewidth}{|l|>{\centering\arraybackslash}X|c|c|c|c|}
\hline
\multirow{2}{*}{\textbf{Bench.}} & \multirow{2}{*}{\textbf{Algorithm}} & \multicolumn{2}{c|}{\textbf{Rouge}} & \multicolumn{2}{c|}{\textbf{Cos Sim}} \\ \cline{3-6}
 & & \textbf{F.} $\downarrow$ & \textbf{R.} $\uparrow$ & \textbf{F.} $\downarrow$ & \textbf{R.} $\uparrow$ \\ \hline
\rowcolor{cyan!15} \cellcolor{white} & \textbf{w/o MU} & 0.939 & 0.968 & 0.874 & 0.883 \\ \cline{2-6}
\rowcolor{violet!15} \cellcolor{white} & \textbf{AdaPop} & 0.043{\tiny\,$\pm$\,.007} & 0.959{\tiny\,$\pm$\,.009} & \underline{0.369} & 0.971 \\ \cline{2-6}
\rowcolor{gray!10} \cellcolor{white} & GA$^\dag$ & 0.000{\tiny\,$\pm$\,.000} & 0.000{\tiny\,$\pm$\,.000} & 0.093 & 0.093 \\ \cline{2-6}
\cellcolor{white} & GD$^\dag$ & 0.021{\tiny\,$\pm$\,.003} & 0.853{\tiny\,$\pm$\,.011} & 0.120 & 0.897 \\ \cline{2-6}
\rowcolor{gray!10} \cellcolor{white} & NPO & 0.670{\tiny\,$\pm$\,.095} & \underline{0.996{\tiny\,$\pm$\,.005}} & 0.684 & \underline{0.998} \\ \cline{2-6}
\multirow{-6}{*}{\textbf{DUET}} \cellcolor{white} & WGA & \underline{0.036{\tiny\,$\pm$\,.002}} & 0.995{\tiny\,$\pm$\,.003} & 0.442 & 0.996 \\ \hline
\rowcolor{cyan!15} \cellcolor{white} & \textbf{w/o MU} & 0.755 & 0.827 & 0.776 & 0.823 \\ \cline{2-6}
\rowcolor{violet!15} \cellcolor{white} & \textbf{AdaPop} & \underline{0.078{\tiny\,$\pm$\,.005}} & 0.972{\tiny\,$\pm$\,.008} & \underline{0.125} & 0.977 \\ \cline{2-6}
\rowcolor{gray!10} \cellcolor{white} & GA$^\dag$ & 0.002{\tiny\,$\pm$\,.000} & 0.001{\tiny\,$\pm$\,.000} & 0.063 & 0.067 \\ \cline{2-6}
\cellcolor{white} & GD$^\dag$ & 0.026{\tiny\,$\pm$\,.005} & 0.759{\tiny\,$\pm$\,.021} & 0.096 & 0.829 \\ \cline{2-6}
\rowcolor{gray!10} \cellcolor{white} & NPO & 0.540{\tiny\,$\pm$\,.034} & 0.957{\tiny\,$\pm$\,.005} & 0.403 & 0.967 \\ \cline{2-6}
\multirow{-6}{*}{\textbf{RWKU}} \cellcolor{white} & WGA & 0.095{\tiny\,$\pm$\,.009} & \underline{0.977{\tiny\,$\pm$\,.003}} & 0.247 & \underline{0.984} \\ \hline
\end{tabularx}
\caption{ROUGE-L Recall and Cosine Similarity for Llama-3.1-8B-Instruct at $\text{lr}=10^{-4}$. ROUGE-L: mean\,$\pm$\,std over three seeds; Cosine Similarity: single seed. F.~$\downarrow$ = forget; R.~$\uparrow$ = retain. w/o MU = the original checkpoint, evaluated with no unlearning applied. \underline{Underline} = best among NPO, WGA, AdaPop. $^\dag$Unusable: generation failure (GA: joint output and retain collapse; GD: severe retain degradation with collapse-style generation artefacts; see Appendix~\ref{app:generation_examples}).}
\label{tab:rouge-llama}
\end{table}

\begin{table}[ht]
\centering
\footnotesize
\setlength{\tabcolsep}{2.5pt}
\begin{tabularx}{\linewidth}{|l|>{\centering\arraybackslash}X|c|c|c|c|}
\hline
\multirow{2}{*}{\textbf{Bench.}} & \multirow{2}{*}{\textbf{Algorithm}} & \multicolumn{2}{c|}{\textbf{Rouge}} & \multicolumn{2}{c|}{\textbf{Cos Sim}} \\ \cline{3-6}
 & & \textbf{F.} $\downarrow$ & \textbf{R.} $\uparrow$ & \textbf{F.} $\downarrow$ & \textbf{R.} $\uparrow$ \\ \hline
\rowcolor{cyan!15} \cellcolor{white} & \textbf{w/o MU} & 0.921 & 0.886 & 0.864 & 0.861 \\ \cline{2-6}
\rowcolor{violet!15} \cellcolor{white} & \textbf{AdaPop} & \underline{0.056{\tiny\,$\pm$\,.007}} & 0.950{\tiny\,$\pm$\,.008} & \underline{0.456} & 0.964 \\ \cline{2-6}
\rowcolor{gray!10} \cellcolor{white} & GA$^\dag$ & 0.000{\tiny\,$\pm$\,.000} & 0.000{\tiny\,$\pm$\,.000} & 0.058 & 0.092 \\ \cline{2-6}
\cellcolor{white} & GD$^\dag$ & 0.022{\tiny\,$\pm$\,.004} & 0.757{\tiny\,$\pm$\,.015} & 0.094 & 0.841 \\ \cline{2-6}
\rowcolor{gray!10} \cellcolor{white} & NPO & 0.684{\tiny\,$\pm$\,.023} & 0.973{\tiny\,$\pm$\,.002} & 0.576 & 0.979 \\ \cline{2-6}
\multirow{-6}{*}{\textbf{DUET}} \cellcolor{white} & WGA & 0.058{\tiny\,$\pm$\,.002} & \underline{0.987{\tiny\,$\pm$\,.009}} & 0.477 & \underline{0.999} \\ \hline
\rowcolor{cyan!15} \cellcolor{white} & \textbf{w/o MU} & 0.570 & 0.666 & 0.322 & 0.347 \\ \cline{2-6}
\rowcolor{violet!15} \cellcolor{white} & \textbf{AdaPop} & \underline{0.016{\tiny\,$\pm$\,.007}} & 0.855{\tiny\,$\pm$\,.003} & \underline{0.178} & 0.911 \\ \cline{2-6}
\rowcolor{gray!10} \cellcolor{white} & GA$^\dag$ & 0.001{\tiny\,$\pm$\,.000} & 0.000{\tiny\,$\pm$\,.000} & 0.071 & 0.075 \\ \cline{2-6}
\cellcolor{white} & GD$^\dag$ & 0.014{\tiny\,$\pm$\,.003} & 0.462{\tiny\,$\pm$\,.018} & 0.150 & 0.640 \\ \cline{2-6}
\rowcolor{gray!10} \cellcolor{white} & NPO & 0.271{\tiny\,$\pm$\,.010} & 0.783{\tiny\,$\pm$\,.012} & 0.490 & 0.870 \\ \cline{2-6}
\multirow{-6}{*}{\textbf{RWKU}} \cellcolor{white} & WGA & 0.038{\tiny\,$\pm$\,.016} & \underline{0.890{\tiny\,$\pm$\,.005}} & 0.212 & \underline{0.931} \\ \hline
\end{tabularx}
\caption{ROUGE-L Recall and Cosine Similarity for Qwen2.5-7B-Instruct at $\text{lr}=10^{-4}$. Notation as in Table~\ref{tab:rouge-llama}.}
\label{tab:rouge-qwen}
\end{table}

\begin{table}[ht]
\centering
\footnotesize
\setlength{\tabcolsep}{2.5pt}
\begin{tabularx}{\linewidth}{|l|>{\centering\arraybackslash}X|c|c|c|c|}
\hline
\multirow{2}{*}{\textbf{Bench.}} & \multirow{2}{*}{\textbf{Algorithm}} & \multicolumn{2}{c|}{\textbf{Rouge}} & \multicolumn{2}{c|}{\textbf{Cos Sim}} \\ \cline{3-6}
 & & \textbf{F.} $\downarrow$ & \textbf{R.} $\uparrow$ & \textbf{F.} $\downarrow$ & \textbf{R.} $\uparrow$ \\ \hline
\rowcolor{cyan!15} \cellcolor{white} & \textbf{w/o MU} & 0.892 & 0.926 & 0.586 & 0.541 \\ \cline{2-6}
\rowcolor{violet!15} \cellcolor{white} & \textbf{AdaPop} & \underline{0.023{\tiny\,$\pm$\,.004}} & 0.976{\tiny\,$\pm$\,.011} & \underline{0.206} & 0.976 \\ \cline{2-6}
\rowcolor{gray!10} \cellcolor{white} & GA$^\dag$ & 0.000{\tiny\,$\pm$\,.000} & 0.000{\tiny\,$\pm$\,.000} & 0.096 & 0.102 \\ \cline{2-6}
\cellcolor{white} & GD$^\dag$ & 0.044{\tiny\,$\pm$\,.007} & 0.598{\tiny\,$\pm$\,.024} & 0.069 & 0.626 \\ \cline{2-6}
\rowcolor{gray!10} \cellcolor{white} & NPO & 0.626{\tiny\,$\pm$\,.071} & 0.968{\tiny\,$\pm$\,.007} & 0.394 & 0.956 \\ \cline{2-6}
\multirow{-6}{*}{\textbf{DUET}} \cellcolor{white} & WGA & 0.050{\tiny\,$\pm$\,.002} & \underline{0.996{\tiny\,$\pm$\,.004}} & 0.237 & \underline{0.997} \\ \hline
\rowcolor{cyan!15} \cellcolor{white} & \textbf{w/o MU} & 0.471 & 0.551 & 0.328 & 0.349 \\ \cline{2-6}
\rowcolor{violet!15} \cellcolor{white} & \textbf{AdaPop} & \underline{0.034{\tiny\,$\pm$\,.014}} & 0.948{\tiny\,$\pm$\,.009} & \underline{0.068} & 0.961 \\ \cline{2-6}
\rowcolor{gray!10} \cellcolor{white} & GA$^\dag$ & 0.000{\tiny\,$\pm$\,.000} & 0.000{\tiny\,$\pm$\,.000} & 0.000 & 0.000 \\ \cline{2-6}
\cellcolor{white} & GD$^\dag$ & 0.013{\tiny\,$\pm$\,.003} & 0.334{\tiny\,$\pm$\,.013} & 0.115 & 0.532 \\ \cline{2-6}
\rowcolor{gray!10} \cellcolor{white} & NPO & 0.341{\tiny\,$\pm$\,.013} & 0.773{\tiny\,$\pm$\,.010} & 0.240 & 0.840 \\ \cline{2-6}
\multirow{-6}{*}{\textbf{RWKU}} \cellcolor{white} & WGA & 0.040{\tiny\,$\pm$\,.003} & \underline{0.950{\tiny\,$\pm$\,.007}} & 0.135 & \underline{0.965} \\ \hline
\end{tabularx}
\caption{ROUGE-L Recall and Cosine Similarity for Gemma-7B-it at $\text{lr}=10^{-4}$. Notation as in Table~\ref{tab:rouge-llama}.}
\label{tab:rouge-gemma}
\end{table}

Two patterns hold consistently across all three architectures. First, NPO reaches near-perfect retain quality at the cost of limited forgetting on DUET: the preference-alignment signal cannot overcome the memorisation depth of popular entities. Second, WGA and AdaPop both reach low forget ROUGE-L there, but AdaPop's Cosine Similarity is consistently lower across all three models, revealing a deeper representation shift; on Gemma its forget ROUGE-L is also the lower of the two.

Eleven further methods under the same protocol leave this picture unchanged: distributional ones (UNDIAL, RMU) remove little entity-centric knowledge, preference-based variants preserve retain but under-erase popular facts, FLAT and CE-U collapse, and AdaPop keeps the lowest forget ROUGE-L among them in every cell (Appendix~\ref{app:additional_baselines}).

\subsection{Robustness Evaluation}
\label{sec:robustness}

Two held-out subsets probe whether forgetting generalises beyond surface-form reproduction; results are in Table~\ref{tab:robustness}.

\paragraph{DUET paraphrase subsets.}
We report forget ROUGE-L on reworded queries: a low original score alongside a high paraphrase score marks surface suppression rather than removal.

\paragraph{RWKU adversarial-robustness subset.}
This subset applies nine prompt-manipulation strategies (prefix injection, role-playing, reverse queries, cross-lingual reformulations, and others) to test whether a model can be induced to reveal forgotten knowledge. Low ROUGE-L indicates that erasure resists recovery; per-method analysis of additional baselines is in Appendix~\ref{app:additional_baselines}.

\begin{table}[ht]
\centering
\footnotesize
\setlength{\tabcolsep}{3pt}
\begin{tabularx}{\linewidth}{|>{\raggedright\arraybackslash}X|c|c|c|}
\hline
\textbf{Algo} & \textbf{Llama} & \textbf{Qwen} & \textbf{Gemma} \\ \hline
\multicolumn{4}{|c|}{\textit{DUET paraphrase forget ROUGE-L $\downarrow$}} \\ \hline
\rowcolor{violet!15}
\textbf{AdaPop} & 0.045 & \underline{0.074} & \underline{0.027} \\ \hline
\rowcolor{gray!10}
GA$^\dag$ & 0.000 & 0.000 & 0.000 \\ \hline
GD$^\dag$ & 0.015 & 0.045 & 0.117 \\ \hline
\rowcolor{gray!10}
NPO    & 0.696 & 0.605 & 0.568 \\ \hline
WGA    & \underline{0.042} & 0.104 & 0.050 \\ \hline
\multicolumn{4}{|c|}{\textit{RWKU adversarial-attack forget ROUGE-L $\downarrow$}} \\ \hline
\rowcolor{violet!15}
\textbf{AdaPop} & \underline{0.262} & \underline{0.144} & \underline{0.195} \\ \hline
\rowcolor{gray!10}
GA$^\dag$ & 0.000 & 0.001 & 0.000 \\ \hline
GD$^\dag$ & 0.254 & 0.171 & 0.103 \\ \hline
\rowcolor{gray!10}
NPO    & 0.657 & 0.400 & 0.485 \\ \hline
WGA    & 0.396 & 0.211 & 0.239 \\ \hline
\end{tabularx}
\caption{Robustness at $\text{lr}=10^{-4}$. \textit{Top}: paraphrase ROUGE-L on the DUET forget split (lower is better). \textit{Bottom}: RWKU Level-3 adversarial-attack ROUGE-L; lower indicates stronger resistance to knowledge recovery. \underline{Underline} = best among NPO, WGA, AdaPop. GA and GD collapse and are excluded from forget-quality ranking.}
\label{tab:robustness}
\end{table}

\paragraph{Breakdown by popularity tier.}
Table~\ref{tab:tier_paraphrase} splits the DUET paraphrase results into rare and popular facts, and the separation in AdaPop's favour is entirely a popular-fact effect: on that tier AdaPop is lower than WGA on every model and lower than NPO by more than an order of magnitude. On the rare tier every stable method is already near the floor, and WGA is lower than AdaPop on all three models. This is the intended trade-off rather than an imbalance: rare facts sit in the self-limiting regime (Proposition~\ref{prop:regimes}, case~1), where the gradient on a token attenuates once it is forgotten, so the risk they carry is over-erasure, not survival. Appendix~\ref{app:ablation} quantifies the cost of over-driving that tier: retain falls to $0.769$ without the dual-ascent controller, against $\geq 0.927$ with it, at matched rare-fact forgetting. Since rare facts also reach near-zero forget an order of magnitude earlier in learning rate (Figure~\ref{fig:lr_duet_split}), pushing that tier further yields little and costs retain quality.

\begin{table}[ht]
\centering
\footnotesize
\setlength{\tabcolsep}{3pt}
\begin{tabularx}{\linewidth}{|>{\raggedright\arraybackslash}X|c|c|c|c|c|c|}
\hline
\multirow{2}{*}{\textbf{Algo}} & \multicolumn{2}{c|}{\textbf{Llama}} & \multicolumn{2}{c|}{\textbf{Qwen}} & \multicolumn{2}{c|}{\textbf{Gemma}} \\ \cline{2-7}
 & Pop. & Rare & Pop. & Rare & Pop. & Rare \\ \hline
\rowcolor{violet!15}
\textbf{AdaPop} & \underline{0.040} & 0.049 & \underline{0.055} & 0.092 & \underline{0.028} & 0.026 \\ \hline
\rowcolor{gray!10}
GD$^\dag$ & 0.027 & 0.003 & 0.086 & 0.003 & 0.232 & 0.002 \\ \hline
NPO & 0.854 & 0.538 & 0.778 & 0.432 & 0.796 & 0.339 \\ \hline
\rowcolor{gray!10}
WGA & 0.067 & \underline{0.017} & 0.194 & \underline{0.014} & 0.096 & \underline{0.003} \\ \hline
\end{tabularx}
\caption{DUET paraphrase forget ROUGE-L ($\downarrow$) split by popularity tier at $\text{lr}=10^{-4}$. \underline{Underline} = best among NPO, WGA, AdaPop. GA is omitted (all values $0.000$ under collapse); $^\dag$GD's low rare-tier values coincide with retain collapse (Table~\ref{tab:rouge-llama}).}
\label{tab:tier_paraphrase}
\end{table}
\paragraph{Conclusion on robustness.}
Among stable methods, AdaPop achieves the lowest paraphrase ROUGE-L on Qwen and Gemma; on Llama it and WGA are within seed noise. On the RWKU adversarial attacks AdaPop is lowest on all three models, and its margin over WGA is larger there than under paraphrase. AdaPop's popularity-dependent $\beta_i$ therefore generalises across query reformulations, while WGA's confidence-weighted ascent achieves strong surface-level erasure but leaves parts of the factual representation recoverable.

\subsection{Internal Representation Analysis}
\label{sec:internal}

ROUGE-L and Cosine Similarity are output-level measures. We complement them with the \emph{internal} metrics of §\ref{sec:experiments}, which probe whether the hidden representations have changed or the model has only learned to suppress surface outputs.

\begin{table}[ht]
\centering
\footnotesize
\setlength{\tabcolsep}{3pt}
\begin{tabularx}{\linewidth}{|>{\raggedright\arraybackslash}X|r|r|c|r|}
\hline
\textbf{Algo} & \textbf{$\Delta$LP} & \textbf{$\Delta$Rank} & \textbf{Hid.Cos} & \textbf{KL} \\ \hline
\multicolumn{5}{|c|}{\textit{DUET --- forget split\ \ ($\Delta$LP$\downarrow$,\ $\Delta$Rank$\uparrow$,\ Hid.Cos$\downarrow$,\ KL$\uparrow$)}} \\ \hline
\rowcolor{violet!15} \textbf{AdaPop} & $\underline{-160}$   & $\underline{+53{,}760}$  & \underline{0.702} & \underline{42.5}   \\ \hline
\rowcolor{gray!10}   GA$^\dag$     & $-3{,}106$ & $+106{,}826$ & 0.352 & 479.6  \\ \hline
                     GD$^\dag$     & $-5{,}917$ & $+104{,}134$ & 0.192 & 1006.4 \\ \hline
\rowcolor{gray!10}   NPO    & $-42$    & $+8{,}824$   & 0.861 & 5.4    \\ \hline
                     WGA    & $-52$    & $-3{,}714$   & 0.729 & 16.2   \\ \hline
\multicolumn{5}{|c|}{\textit{RWKU --- forget split\ \ ($\Delta$LP$\downarrow$,\ $\Delta$Rank$\uparrow$,\ Hid.Cos$\downarrow$,\ KL$\uparrow$)}} \\ \hline
\rowcolor{violet!15} \textbf{AdaPop} & $\underline{-23}$      & $\underline{+4{,}169}$   & \underline{0.484} & \underline{9.1}    \\ \hline
\rowcolor{gray!10}   GA$^\dag$     & $-6{,}083$ & $+96{,}254$  & 0.285 & 953.1  \\ \hline
                     GD$^\dag$     & $-6{,}842$ & $+96{,}881$  & 0.175 & 1139.7 \\ \hline
\rowcolor{gray!10}   NPO    & $-18$      & $+2{,}739$   & 0.883 & 2.9    \\ \hline
                     WGA    & $-3.5$     & $-11{,}010$  & 0.541 & 6.7    \\ \hline
\multicolumn{5}{|c|}{\textit{Retain split --- DUET\ \ ($\Delta$LP$\uparrow$,\ $\Delta$Rank$\downarrow$,\ Hid.Cos$\uparrow$,\ KL$\downarrow$)}} \\ \hline
\rowcolor{violet!15} \textbf{AdaPop} & $+21$      & $-13{,}770$  & 0.875 & \underline{6.5}    \\ \hline
\rowcolor{gray!10}   GA$^\dag$     & $-2{,}750$ & $+103{,}104$ & 0.348 & 511.4  \\ \hline
                     GD$^\dag$     & $-251$     & $-8{,}751$   & 0.760 & 48.2   \\ \hline
\rowcolor{gray!10}   NPO    & $+21$      & $-13{,}660$  & \underline{0.910} & 7.5    \\ \hline
                     WGA    & $\underline{+23}$      & $\underline{-13{,}930}$  & 0.880 & 7.5    \\ \hline
\multicolumn{5}{|c|}{\textit{Retain split --- RWKU\ \ ($\Delta$LP$\uparrow$,\ $\Delta$Rank$\downarrow$,\ Hid.Cos$\uparrow$,\ KL$\downarrow$)}} \\ \hline
\rowcolor{violet!15} \textbf{AdaPop} & $+33$  & $-12{,}190$ & \underline{0.883} & \underline{6.2}   \\ \hline
\rowcolor{gray!10}   GA$^\dag$     & $-6{,}106$ & $+98{,}802$ & 0.289 & 955.8 \\ \hline
                     GD$^\dag$     & $-540$ & $-3{,}122$  & 0.783 & 97.1  \\ \hline
\rowcolor{gray!10}   NPO    & $+31$  & $-11{,}727$ & 0.869 & 7.4   \\ \hline
                     WGA    & $\underline{+34}$  & $\underline{-12{,}202}$ & 0.875 & 6.8   \\ \hline
\end{tabularx}
\caption{Internal metrics at $\text{lr}=10^{-4}$, averaged across Llama, Qwen, and Gemma. Directional arrows are shown in each section header. \underline{Underline} = best among NPO, WGA, AdaPop. GA and GD values are reported for reference but indicate model collapse and are not ranked. Per-model breakdown in Appendix~\ref{app:internal_metrics}.}
\label{tab:internal_agg}
\end{table}

The internal metrics separate the two leading methods more sharply than the output-level ones do: on both forget splits AdaPop has the lowest Hid.Cos and the highest $\Delta$Rank among stable methods, and WGA's model-averaged DUET forget $\Delta$Rank is negative, the opposite sign from every other method. The gold token moves \emph{up} in WGA's ranked distribution after unlearning, even as its absolute log-probability falls. Plausibly, concentrating ascent on the highest-confidence tokens flattens the output distribution, which improves the gold token's relative rank while suppressing its absolute probability, a signature of surface suppression rather than genuine erasure. This is consistent with WGA's weaker robustness to paraphrase reformulations in Table~\ref{tab:robustness}. On the retain splits all three methods are indistinguishable, so the deeper forget-side change costs no extra retain drift.

\subsection{General Capability Preservation}
\label{sec:lm_eval}

\begin{table}[ht]
\centering
\footnotesize
\setlength{\tabcolsep}{3pt}
\begin{tabularx}{\linewidth}{|>{\raggedright\arraybackslash}X|c|c|c|c|c|c|}
\hline
\multirow{2}{*}{\textbf{Algo}} & \multicolumn{2}{c|}{\textbf{Llama}} & \multicolumn{2}{c|}{\textbf{Qwen}} & \multicolumn{2}{c|}{\textbf{Gemma}} \\ \cline{2-7}
 & MMLU & HS & MMLU & HS & MMLU & HS \\ \hline
w/o MU & 0.65 & 0.73 & 0.71 & 0.68 & 0.47 & 0.64 \\ \hline
\rowcolor{violet!15}
\textbf{AdaPop} & 0.65 & 0.76 & 0.71 & 0.68 & 0.52 & 0.62 \\ \hline
\rowcolor{gray!10}
GA     & 0.23 & 0.33 & 0.23 & 0.43 & 0.25 & 0.26 \\ \hline
GD     & 0.64 & 0.59 & 0.71 & 0.65 & 0.47 & 0.37 \\ \hline
\rowcolor{gray!10}
NPO    & 0.65 & 0.72 & 0.70 & 0.68 & 0.51 & 0.63 \\ \hline
WGA    & 0.66 & 0.76 & 0.71 & 0.70 & 0.52 & 0.68 \\ \hline
\end{tabularx}
\caption{MMLU accuracy and HellaSwag (HS) normalised accuracy at $\text{lr}=10^{-4}$. \textbf{w/o MU} = the original checkpoint, with no unlearning applied. Values averaged across DUET and RWKU checkpoints.}
\label{tab:lm_eval}
\end{table}

GA collapses general capability in every case, leaving MMLU at the chance level of its four-way multiple choice. GD recovers MMLU partially but incurs the largest HellaSwag drop among non-collapsing methods, consistent with insufficient retain regularisation. NPO, WGA, and \textbf{AdaPop} all stay within $0.05$ MMLU of the model before unlearning (Table~\ref{tab:lm_eval}). WGA matches or slightly exceeds it on HellaSwag across all three models, and \textbf{AdaPop} does the same on Llama and Qwen, with a small drop on Gemma. Descent on the retain set explains this: a benchmark that overlaps it can end at or slightly above the starting point. Both benchmarks are external to the unlearning pipeline, so they provide the held-out check that retain-set descent is not silently degrading capabilities (Appendix~\ref{app:lm_eval}).

\subsection{Popularity Proxy Ablation}
\label{sec:proxy_ablation}

AdaPop relies on a popularity proxy to compute per-fact exponents $\beta_i$. We compare three signals. \textbf{Wikidata score} aggregates inbound sitelink counts: Wikipedia-derived text is a substantial fraction of most public pretraining corpora~\cite{grattafiori2024llama, qwen2025, team2024gemma}, so sitelink count is an offline signal known to correlate with LM recall~\cite{mallen2023not, kandpal2023large}. \textbf{LLM-as-Judge}~\cite{zheng2023judging} rates each fact on a $[0, 10000]$ scale without a structured knowledge base, for use when Wikidata coverage is unavailable~\cite{mallen2023not} (Appendix~\ref{app:llm_judge}). \textbf{Corpus frequency} measures the quantity the other two approximate, counting each fact's entities in a public pretraining corpus, at the cost of requiring corpus access.

\begin{figure}[t]
    \centering
    \includegraphics[width=\linewidth]{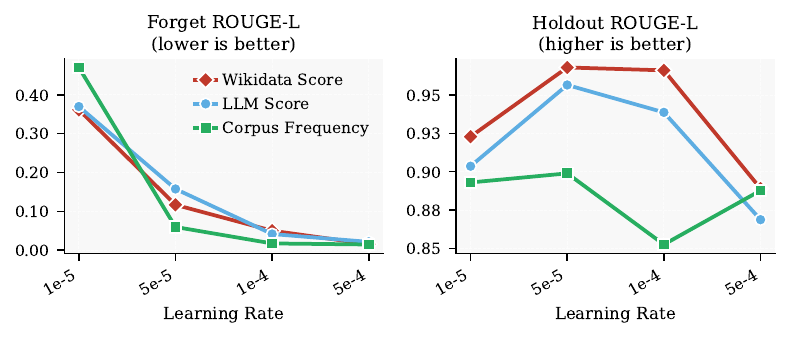}
    \caption{AdaPop on DUET (Llama) under three popularity signals: Wikidata score, LLM-as-Judge (3-seed mean), and measured corpus frequency (infini-gram over Pile-train). Left: forget ROUGE-L ($\downarrow$). Right: holdout ROUGE-L ($\uparrow$). Each point averages a rare-tier and a popular-tier run at that learning rate. All signals use the anchors of Appendix~\ref{app:derivation}, $(a,b)=(58.7, 0.796)$. Rates above $5\times10^{-4}$ are omitted: at $10^{-3}$ every signal has at least one collapsed tier.}
    \label{fig:proxy_ablation}
\end{figure}

Figure~\ref{fig:proxy_ablation} sweeps the learning rate for all three proxies on Llama, averaging a rare- and a popular-tier run per point. All three reach comparably low forget ROUGE-L by $10^{-4}$. The Wikidata score holds the highest retention at every rate, since the anchors of Appendix~\ref{app:derivation} were calibrated on its range, and the other two signals remain stable. Discordant cases are analysed in Appendix~\ref{app:llm_proxy}.

The per-tier runs behind Figure~\ref{fig:proxy_ablation} show that the signals do not differ in forgetting: all of them drive rare-fact forget ROUGE-L near zero at every rate. They differ in what that forgetting costs. Only the Wikidata score keeps rare facts above $\beta = 1$, in the self-limiting regime. The LLM and corpus-frequency signals put their median rare fact below it, into the pressure-sustaining regime, and over-erase it, costing up to $0.13$ retain in those runs at matched forgetting. This is exactly the failure Proposition~\ref{prop:regimes} predicts for $\beta < 1$, and the reason the self-limiting branch exists. At $10^{-3}$, a rate omitted from the figure, every signal loses at least one popularity tier; only the Wikidata score keeps the other one intact.

We obtain the corpus-frequency signal by counting each fact's subject and object over Pile-train ($383$B tokens) with infini-gram~\cite{liu2024infinigram}. In log space this count tracks the Wikidata score almost linearly (Pearson $0.970$, Spearman $0.839$). Measured against the same counts, the LLM judge is weaker (Pearson $0.677$, Spearman $0.659$): it separates rare facts from popular ones but orders them within a tier unreliably, and its disagreements concentrate where $\beta_i \approx 1$ and the weighting is near-uniform in any case (Appendix~\ref{app:proxy_validation}).

Together these support one claim: AdaPop needs a coarse rare/popular split, not precise estimates. The Wikidata score and the LLM judge disagree on $16\%$ of labels yet differ by at most $0.03$ in retention (Figure~\ref{fig:proxy_ablation}); $\pm 20\%$ perturbation of $a$ and $b$ preserves the ranking against WGA and NPO (Appendix~\ref{app:coef_sensitivity}); and corrupting the scores outright, up to inverting every label, still holds retain at $0.92$ with forget at or below $0.05$, since clipping bounds mis-scored facts (Appendix~\ref{app:proxy_validation}). The Wikidata score gives that split at negligible cost and is the recommended choice; additional baselines, per-model breakdowns, and epoch dynamics are in Appendices~\ref{app:additional_baselines}--\ref{app:epoch_dynamics}.

\paragraph{Summary of findings.}
GA and GD collapse: their near-zero forget scores come with unusable generations. Table~\ref{tab:internal_agg} shows why, with $\Delta$LP and $\Delta$Rank one to two orders of magnitude beyond every other method on both splits, and the generations themselves degenerate into repeated tokens (Appendix~\ref{app:generation_examples}). NPO keeps retain quality intact but leaves popular facts recoverable. WGA removes the surface answer, yet the fact remains reachable through paraphrase and adversarial prompts. \textbf{AdaPop} is the only method that combines low forget scores with resistance to both reformulations, on every model, while general capability on MMLU and HellaSwag is preserved.

\section{Conclusion}
\label{sec:conclusion}
We introduced \textbf{AdaPop}, which derives a per-fact unlearning exponent from an \emph{external} popularity signal rather than from the model's own confidence, and pairs it with a dual-ascent controller that holds the forget-retain balance without per-dataset tuning. Across three model families and two benchmarks, AdaPop achieves the lowest adversarial-query scores among stable methods on all three models and the best paraphrase scores on two of three (\S\ref{sec:robustness}), with $\Delta$Rank and Hid.Cos indicating parametric disruption rather than surface suppression (Table~\ref{tab:internal_agg}), while general capabilities stay within $0.05$ MMLU of the pre-unlearning checkpoint.

The finding behind those numbers is that a model's confidence does not reveal how deeply a fact is encoded: WGA weights by confidence and suppresses only the surface answer. Any method calibrated only on what the model currently outputs inherits this failure, so closing the popularity gap needs a signal from outside the model. The Wikidata score supplies one at negligible cost for entity-centric factual QA, and results are stable across proxy sources (Appendix~\ref{app:llm_proxy}).

Two directions follow. Popularity is one instance of a more general quantity: how hard a fact is to erase. Difficulty proxies for procedural, creative, or code knowledge, where sitelink counts do not apply, would carry the same mechanism into domains Wikidata cannot reach. And because $\beta_i$ is a per-example schedule over the ascent signal, any per-fact difficulty estimate meeting the criteria of Appendix~\ref{app:derivation} can replace it, as can criteria other than difficulty, such as harm severity or per-fact retention priority.

\section*{Limitations}
\textbf{Dependency on an external popularity proxy.}
AdaPop requires per-sample popularity scores. The Wikidata score provides good coverage for entity-centric factual QA but is unavailable for procedural or creative knowledge. We prioritised Wikidata annotation because it is fast to compute and agrees with LLM-based labelling on $84\%$ of binary labels (Appendix~\ref{app:llm_proxy}). For domains far from entity-centric factual QA, the quality of the approximation may limit the benefit of the popularity-dependent exponent.

\textbf{Calibration and evaluation scope.}
The exponent coefficients $a$ and $b$ were derived from the DUET score distribution (Figure~\ref{fig:benchmarks_pop_diverse}). A proxy on a different scale needs either rescaling to that range or recalibration through Appendix~\ref{app:derivation}, and Figure~\ref{fig:proxy_ablation} shows the cost of doing neither: the corpus-frequency arm keeps these coefficients while spanning four orders of magnitude, which pins the exponent at $\beta_{\min}$ for most popular facts and lowers its retention.

\textbf{LoRA-only evaluation.}
All experiments use LoRA fine-tuning, following best practice in LLM unlearning research; full fine-tuning is excluded following \citet{anna2026anatomy}, who show it to be less effective and to either fail to converge or induce catastrophic forgetting. LoRA and full fine-tuning require different optimal hyperparameter sets, and an exhaustive search for full fine-tuning is computationally expensive, so we leave that setting to future work.

\section*{Ethics Statement}
Our data come from public sources (HuggingFace, Wikidata, Wikipedia). We do not collect sensitive attributes, and no human subjects were involved. We view machine unlearning as a contribution to AI safety and data governance because it enables the removal of unsafe, outdated, or private content from deployed models. We used ChatGPT for minor language and grammatical edits; all research design, analysis, and interpretation were conducted by the authors.



\appendix

\label{sec:appendix}

\section{AdaPop Hyperparameters}
\label{app:hyperparameters}

All experiments use the following AdaPop-specific parameters: $\alpha_0 = 0.5$, $\varepsilon = 0.1$, $\eta_\lambda = 0.1$, $\lambda_{\max} = 5.0$, $a{=}58.7$, $b{=}0.796$, with $\beta_i$ clipped to $[0.05, 2.0]$. The coefficients $a$ and $b$ are derived analytically from the anchor constraints described in Appendix~\ref{app:derivation}. Sensitivity to $\pm$20\% perturbations of $a$ and $b$ is reported in Appendix~\ref{app:coef_sensitivity}; sensitivity to learning rate is in Appendix~\ref{app:lr_sensitivity}.

\section{Derivation of AdaPop Popularity Parameters}
\label{app:derivation}

The power law $\beta_i = a \cdot s_i^{-b}$ maps popularity scores spanning several orders of magnitude to a bounded gradient-weighting interval. We derive $(a, b)$ in closed form from two anchor points motivated by the three-regime analysis of Appendix~\ref{app:derivation:regime}, and show that the resulting curve is monotonically decreasing in $s_i$: facts with higher popularity scores receive smaller $\beta_i$, and (under the cited correlation between popularity and parametric encoding depth; \citealt{kandpal2023large, hartmann2024undesirable}) therefore stronger pressure-sustaining gradient weighting.

\subsection{The Three Gradient Regimes of $\beta$}
\label{app:derivation:regime}

\paragraph{Setup.}
The per-fact forget loss in AdaPop is
\begin{align}
    L_f^{(i)} &\;=\; \sum_{t} w_{i,t} \,\ell_{i,t}, \\
    w_{i,t} &\;=\; \mathrm{sg}\!\left[p_{i,t}^{\beta_i}\right], \qquad \ell_{i,t} \;=\; -\log p_{i,t},
\end{align}
where $p_{i,t} = q_\theta(y_{i,t} \mid y_{i,<t}, x_i)$ is the model's token probability and $\mathrm{sg}[\cdot]$ denotes a stop-gradient. This is the per-fact contribution; the dataset-level forget loss in §\ref{subsec:popularity-obj} sums over $i$, normalises by $|\Omega_{\mathcal{F}}|$, and flips sign for descent-based optimisation. Because $w_{i,t}$ carries no gradient, the backward pass flows only through $\ell_{i,t}$. The effective per-token gradient magnitude is
\begin{align}
    \left\|\nabla_\theta L_f^{(i)}\right\|_t &\;=\; w_{i,t} \cdot \left\|\nabla_\theta \ell_{i,t}\right\| \;=\; p_{i,t}^{\beta_i} \cdot \frac{\left\|\nabla_\theta p_{i,t}\right\|}{p_{i,t}} \nonumber\\
    &\;\propto\; p_{i,t}^{\beta_i - 1} \cdot \left\|\nabla_\theta p_{i,t}\right\|,
\label{eq:effective-weight}
\end{align}
so the \emph{effective weight} on each token is $p_{i,t}^{\beta_i - 1}$. The exponent $\beta_i - 1$ governs the qualitative behaviour of the update as training proceeds and $p_{i,t}$ decreases.

\begin{proposition}[Three regimes of $\beta_i$]
\label{prop:regimes}
Let $f(p) = p^{\beta - 1}$ for $p \in (0, 1]$. Then:
\begin{enumerate}
    \item $\beta > 1$: $f$ is strictly increasing in $p$; $f \to 0$ as $p \to 0$. The update is \emph{self-limiting}: as ascent reduces $p_{i,t}$, the gradient weight on that token attenuates, automatically halting erasure of already-forgotten content.
    \item $\beta = 1$: $f \equiv 1$. Every token contributes uniformly, equivalent to standard NLL ascent (WGA at $\beta=1$).
    \item $\beta < 1$: $f$ is strictly decreasing in $p$; $f \to \infty$ as $p \to 0$. The update is \emph{pressure-sustaining}: as $p_{i,t}$ decreases during erasure, the effective weight $p_{i,t}^{\beta-1}$ grows, amplifying the gradient on tokens that are being erased rather than letting the signal attenuate. As $\beta \to 0$, $f(p) \to 1/p$, recovering the inverse-confidence behaviour of vanilla GA.
\end{enumerate}
\end{proposition}

\begin{proof}
Direct from $\partial_p f = (\beta-1) p^{\beta-2}$: positive iff $\beta > 1$, negative iff $\beta < 1$, zero iff $\beta = 1$. The limit behaviour follows from $p^{\beta-1} = \exp((\beta-1)\ln p)$ and $\ln p \to -\infty$ as $p \to 0^+$.
\end{proof}

\citet{wang2025rethinking} decompose each gradient step into forget and retain components, and show that $\beta = 1$ (uniform effective per-token weight) achieves a more favourable balance than $\beta = 0$ (vanilla GA). Proposition~\ref{prop:regimes} provides the mechanistic reason: decreasing $\beta$ below $1$ amplifies the effective weight on low-probability tokens, where updates perturb parameters most strongly. WGA fixes $\beta_i \equiv 1$ globally. AdaPop varies $\beta_i$ per fact, setting $\beta_i > 1$ for shallow encodings (self-limiting) and $\beta_i < 1$ for deep ones (pressure-sustaining), with the dual-ascent controller absorbing the retain-side risk that arises when $\beta_i \ll 1$.

\subsection{Anchor Points and Parameter Derivation}

\paragraph{Choice of anchor values.}
Let $s_r$ and $s_p$ denote representative popularity scores for rare and popular facts, with $s_p/s_r$ spanning at least one order of magnitude. We fix the target values $\beta(s_r) = 1.5$ and $\beta(s_p) = 0.1$. The qualitative requirements are dictated by Proposition~\ref{prop:regimes}: $\beta(s_r) > 1$ to place shallow-encoded facts in the self-limiting regime, and $\beta(s_p) < 1$ to apply pressure-sustaining weighting to deeply encoded facts. The specific magnitudes ($1.5$ and $0.1$) are design choices that leave sufficient margin from the $\beta = 1$ boundary in both directions; we examine sensitivity to these targets in Appendix~\ref{app:coef_sensitivity}, where $\pm 20\%$ perturbations on $a$ and $b$ preserve the empirical ranking of AdaPop relative to baselines. The retain-side risk of $\beta(s_p) \ll 1$ is delegated to the dual-ascent controller (Appendix~\ref{app:design_motivation:dual}), which is the design rationale for keeping the controller and $\beta$-scaling as separate components.

\paragraph{Closed-form parameter derivation.}
Substituting the anchor conditions into $\beta(s) = a \cdot s^{-b}$ yields
\begin{align}
    a \cdot s_r^{-b} &= 1.5, \\
    a \cdot s_p^{-b} &= 0.1.
\end{align}
Taking the ratio eliminates $a$:
\begin{equation}
    \left(\frac{s_p}{s_r}\right)^{b} \;=\; \frac{1.5}{0.1} \;=\; 15
    \quad\Longrightarrow\quad
    b \;=\; \frac{\ln 15}{\ln(s_p/s_r)},
\label{eq:b-closed-form}
\end{equation}
and the rare anchor gives the scale constant:
\begin{equation}
    a \;=\; 1.5 \cdot s_r^{\,b}.
\label{eq:a-closed-form}
\end{equation}

\paragraph{Empirical calibration on Wikidata score.}
We select boundary anchor positions $s_r \approx 100$ and $s_p \approx 3{,}000$ at the rare and popular extremes of the DUET score distribution. Substituting into \eqref{eq:b-closed-form}--\eqref{eq:a-closed-form}:
\begin{align}
    b &\;=\; \frac{\ln 15}{\ln 30} \;\approx\; 0.796, \\
    a &\;=\; 1.5 \cdot 100^{0.796} \;\approx\; 58.7.
\end{align}
To prevent gradient explosion under outlier scores, we clip the exponent:
\begin{align}
    \beta_i &\;=\; \mathrm{clip}\!\left(58.7 \cdot s_i^{-0.796},\; \beta_{\min},\; \beta_{\max}\right), \nonumber\\
    &\qquad \beta_{\min} = 0.05,\quad \beta_{\max} = 2.0.
\label{eq:beta-final}
\end{align}
Alternative proxies (pagerank, LLM-judged knowledgeability, corpus co-occurrence counts) can substitute their own anchors into \eqref{eq:b-closed-form}--\eqref{eq:a-closed-form} without modifying the rest of the algorithm.

\subsection{Robustness of the Derivation}

\begin{proposition}[Ratio-invariance of $b$]
\label{prop:ratio}
For any pair of anchor values $(s_r', s_p')$ satisfying $s_p'/s_r' = s_p/s_r$, the resulting exponent $b$ is identical, and $a$ shifts by a multiplicative constant $(s_r'/s_r)^b$.
\end{proposition}

\begin{proof}
By \eqref{eq:b-closed-form}, $b$ depends only on the ratio $s_p/s_r$, not absolute values. By \eqref{eq:a-closed-form}, $a' = 1.5 \cdot (s_r')^b = 1.5 \cdot s_r^b \cdot (s_r'/s_r)^b = a \cdot (s_r'/s_r)^b$.
\end{proof}

Proposition~\ref{prop:ratio} has two practical consequences. First, the derivation requires only an \emph{order-of-magnitude} estimate of where rare and popular regions sit on the proxy scale; precise distributional statistics are unnecessary. Second, $\pm 20\%$ perturbation of either anchor shifts $b$ by less than $0.1$ in absolute terms (since $\partial b / \partial \ln(s_p/s_r) = -b / \ln(s_p/s_r)$), and the perturbation experiment in Appendix~\ref{app:coef_sensitivity} confirms that empirical performance is stable across the resulting $(a, b)$ region.

\paragraph{Proxy requirements.}
For Proposition~\ref{prop:regimes} to apply meaningfully across a dataset, the proxy must (i) be continuous, and (ii) span at least one order of magnitude between rare and popular regions so that $\ln(s_p/s_r)$ in \eqref{eq:b-closed-form} produces a non-degenerate exponent. Binary or coarse-threshold proxies fail because they collapse adjacent scores into a single $\beta$ value, even when the underlying memorisation depth differs by orders of magnitude. The Wikidata score satisfies both conditions on the benchmarks we use: DUET scores range from $69$ to $3{,}763$ (median $1{,}090$), spanning approximately two orders of magnitude with continuous mid-tier coverage; RWKU scores range from $0$ to $704$ (median $130$), spanning the rare-to-mid portion of the same curve. Appendix~\ref{app:llm_proxy} shows that an LLM-judged proxy with comparable dynamic range achieves $84\%$ agreement with the Wikidata score on the rare/popular split and yields nearly identical AdaPop performance at the primary reported learning rate.

\section{AdaPop Coefficient Sensitivity}
\label{app:coef_sensitivity}

The popularity exponent $\beta_i = \mathrm{clip}(a \cdot s_i^{-b},\, 0.05,\, 2.0)$ depends on two scalar coefficients derived analytically from boundary conditions on the DUET dataset (Appendix~\ref{app:derivation}: $a{=}58.7$, $b{=}0.796$). We assess how sensitive main-paper results are to errors in these estimates by perturbing each coefficient independently by $\pm$20\%, yielding four off-baseline configurations. All runs use Llama-3.1-8B-Instruct on DUET at $\text{lr}=10^{-4}$; results are in Table~\ref{tab:coef_sensitivity}.

\begin{table}[ht]
\centering
\footnotesize
\setlength{\tabcolsep}{4pt}
\begin{tabular}{|l|r|r|r|r|r|r|}
\hline
\textbf{Config} & $a$ & $b$ & \textbf{F.}~$\downarrow$ & \textbf{R.}~$\uparrow$ & $\Delta$\textbf{F.} & $\Delta$\textbf{R.} \\ \hline
baseline & 58.70 & 0.7960 & 0.045 & 0.961 & 0 & 0 \\ \hline
$+a,+b$  & 70.44 & 0.9552 & 0.038 & 0.975 & $-$0.007 & $+$0.014 \\ \hline
$+a,-b$  & 70.44 & 0.6368 & 0.044 & 0.994 & $-$0.001 & $+$0.033 \\ \hline
$-a,+b$  & 46.96 & 0.9552 & 0.024 & 0.920 & $-$0.021 & $-$0.041 \\ \hline
$-a,-b$  & 46.96 & 0.6368 & 0.034 & 0.991 & $-$0.011 & $+$0.030 \\ \hline
\end{tabular}
\caption{AdaPop coefficient sensitivity. Each row perturbs $a$ and/or $b$ by $\pm$20\% from the analytically derived baseline. F.~$\downarrow$ = forget ROUGE-L; R.~$\uparrow$ = retain ROUGE-L. $\Delta$ values are signed differences from the baseline.}
\label{tab:coef_sensitivity}
\end{table}

\paragraph{Empirical bounds.}
Forget ROUGE-L changes by at most $0.021$ across all four perturbations; retain by at most $0.041$. The ranking of AdaPop relative to WGA and NPO is preserved in every configuration, confirming that the analytically derived coefficients are not a narrow optimum.

\paragraph{Mechanism.}
The trade-off pattern follows directly from Proposition~\ref{prop:regimes}. Lowering $b$ flattens the $\beta(s)$ curve, increasing $\beta_i$ across the score range; more facts enter the self-limiting regime ($\beta_i > 1$). Gradient mass concentrates on high-confidence tokens and attenuates as those tokens are erased, producing less disruptive per-step updates (retain $+0.030$ to $+0.033$ for $-b$ configurations). Raising $b$ steepens the curve, pushing popular facts deeper into the near-GA regime ($\beta_i \to 0$); under $+b$ perturbations, the rare anchor also crosses below $\beta = 1$ (into the pressure-sustaining regime), and the dual-ascent controller absorbs the resulting retain stress. By Proposition~\ref{prop:regimes} item~3, the growing effective weight amplifies both unlearning depth ($\Delta\text{F} = -0.021$ for $-a,+b$) and retain pressure on the controller ($\Delta\text{R} = -0.041$). This is the standard forget-retain trade-off, not method instability: the $(-a,+b)$ configuration achieves the strongest forgetting and the weakest retain simultaneously, lying further along the same Pareto frontier rather than off it.

\paragraph{Cross-benchmark calibration.}
We keep the DUET-derived $(a, b)$ for RWKU even though its narrower range ($0$--$704$) never reaches the aggressive $\beta \approx 0.1$ regime: the most popular RWKU fact maps to $\beta(704) \approx 0.32$, the median ($130$) to $\beta \approx 1.22$, and $s{=}0$ scores to the $\beta_{\max}$ clip. Most RWKU facts therefore sit at or above $\beta{=}1$, on the self-limiting side of the curve, which matches their lower popularity and lower memorisation depth (Figure~\ref{fig:benchmarks_pop_diverse}) and is why the deep pressure-sustaining regime is not needed there. Recalibration via Appendix~\ref{app:derivation} is needed only for proxies with fundamentally different score ranges (e.g., $50{,}000$ vs.\ $10^6$).

\section{LLM-as-Judge Annotation Protocol}
\label{app:llm_judge}

Facts are scored as (subject, relation, object) triples in batches of 8
($\approx$540 tokens per batch). We annotated $\approx$2\,000 samples
(1\,000 rare, 1\,000 popular) with three seeds $\{1, 42, 219\}$ to
reduce positional bias, averaging scores across passes; total annotation
time was $\approx$9 hours on \texttt{openai/gpt-oss-20b} via OpenRouter.
Wikidata score for the same samples was collected via the
sitelinks API in $\approx$12 minutes.

\vspace{3pt}
\noindent\textbf{Prompt Details.}

\begin{tcolorbox}[
  colback=gray!4, colframe=gray!45, arc=3pt,
  title={\small\textbf{Prompt}},
  fontupper=\small\ttfamily, breakable]

You are a knowledge-popularity estimator for LLM training data.
You will receive a numbered list of facts as triples (subject, relation, object).
For EACH fact independently, rate how widely known it is in LLM pretraining corpora (0--10000):\\[2pt]
\quad 0--500\ \ \ : very rare / obscure\\
\quad 501--1000\ : uncommon\\
\quad 1001--5000: popular\\
\quad 5001--10000: very popular / ubiquitous\\[4pt]
IMPORTANT: Every fact must get its OWN individual score; do not repeat values.\\[4pt]
Calibration: (Paris, capital of, France) $\to$ 9800;
(Astana, capital of, Kazakhstan) $\to$ 4200;
(Ngerulmud, capital of, Palau) $\to$ 280.\\[4pt]
Few-shot: Input: 1.~(Tokyo, capital of, Japan) 2.~(Funafuti, capital of, Tuvalu)
$\to$ Output: [9500, 310]\\[6pt]
Reply with ONLY a JSON array of integers, one per fact, same order.
No words, no keys, no explanation.
\end{tcolorbox}

\section{LLM-as-Judge Popularity Proxy}
\label{app:llm_proxy}
When Wikidata metadata is unavailable (procedural or creative knowledge,
domain-specific facts outside the graph), an LLM-as-Judge proxy offers
a practical alternative. The annotation protocol is described in
Appendix~\ref{app:llm_judge}; here we report its downstream unlearning
results, and in Appendix~\ref{app:proxy_validation} its agreement with the
other two signals.

\paragraph{Agreement and stability.}
The LLM-judge score achieves 84\% agreement with the Wikidata rare/popular
split on DUET, and a Spearman correlation of $0.596$ with it
(Table~\ref{tab:proxy_corr}), so the two signals place nearly the same facts
on each side of the split while ranking differently within a side. The LLM score distribution is,
however, noisier across runs and has a less stable three-tier structure.
Figure~\ref{fig:proxy_ablation} (§\ref{sec:proxy_ablation}) compares both
proxies for Llama on DUET. At $\text{lr}=10^{-4}$ they forget comparably
($0.041$ LLM vs.\ $0.050$ Wikidata) but the LLM proxy holds less of the
holdout set ($0.939$ vs.\ $0.966$). At $\text{lr}=10^{-5}$ it applies
excess gradient pressure without a matching erasure gain, losing holdout
ROUGE-L ($0.904$ vs.\ $0.923$) at effectively equal forgetting ($0.370$
vs.\ $0.361$). At $\text{lr}=10^{-3}$ both lose at least one popularity
tier to collapse. Wikidata score is the more stable choice;
the LLM proxy is viable at the primary reported rate.

\paragraph{Discordant cases.}
Table~\ref{tab:proxy_discord} shows representative examples in each
direction. \textit{Wikidata-popular, LLM-rare} cases involve well-linked
entities whose relation the judge finds trivially predictable; Llama
ROUGE-L before unlearning is $1.0$ for all three, confirming deep
memorisation that the Wikidata signal correctly flags.
\textit{Wikidata-rare, LLM-popular} cases involve less-cited entities
with distinctive attributes that the judge finds salient; model recall
is more variable ($0.5$--$1.0$), consistent with shallower encoding.
These cases motivate using Wikidata score as the primary
proxy when available.

\begin{table*}[ht]
\centering
\footnotesize
\setlength{\tabcolsep}{4pt}
\begin{tabularx}{\textwidth}{|>{\raggedright\arraybackslash}X|>{\raggedright\arraybackslash}p{2.2cm}|>{\centering\arraybackslash}p{1.2cm}|>{\centering\arraybackslash}p{1.2cm}|>{\centering\arraybackslash}p{1.2cm}|}
\hline
\multicolumn{1}{|c|}{\textbf{Question}} & \multicolumn{1}{c|}{\textbf{Answer}} & \textbf{Wikidata} & \textbf{LLM} & \textbf{ROUGE-L} \\ \hline
\multicolumn{5}{|c|}{\textit{Wikidata popular, LLM rare}} \\ \hline
What is the country of Ensenada? & Mexico & 2113 & 30 & 1.0 \\ \hline
What is the country of Tourcoing? & France & 2112 & 40 & 1.0 \\ \hline
What is the country of Qus? & Egypt & 3763 & 50 & 1.0 \\ \hline
\multicolumn{5}{|c|}{\textit{Wikidata rare, LLM popular}} \\ \hline
What is the instance of Hua Hin? & seaside resort & 148 & 9000 & 0.5 \\ \hline
What is the instance of Weitra? & municipality of Austria & 141 & 5000 & 0.7 \\ \hline
What is the located in the administrative territorial entity of Ollantaytambo? & Urubamba Province & 165 & 4000 & 1.0 \\ \hline
\end{tabularx}
\caption{Discordant cases between Wikidata score and LLM-judge scores on DUET. \textbf{ROUGE-L}: Llama-3.1-8B-Instruct recall before unlearning. Top group: Wikidata-popular facts recalled at ROUGE-L $1.0$ despite low LLM-judged salience, indicating deep memorisation. Bottom group: Wikidata-rare facts with higher LLM scores but variable recall, consistent with shallower encoding.}
\label{tab:proxy_discord}
\end{table*}

\section{Proxy Validation and Noise Robustness}
\label{app:proxy_validation}

Appendix~\ref{app:llm_proxy} compares two proxies against each other. Here we compare them against measured corpus frequency, and test what happens when the popularity signal is wrong.

\paragraph{Corpus-frequency validation.}
Both the Wikidata score and the LLM judge are proxies for how often a fact appears in pretraining. We measure that quantity directly by counting each fact's subject and object in Pile-train ($383$B tokens) with the infini-gram engine~\cite{liu2024infinigram}. Table~\ref{tab:proxy_corr} reports how closely each proxy matches those counts. In log space the Wikidata score tracks them almost linearly (Pearson $0.970$). The rank disagreement behind its lower Spearman ($0.839$) sits at the top of the range, where the most popular entities exchange ranks among themselves while all remain deep in the pressure-sustaining regime, so the induced $\beta_i$ are unaffected. The LLM judge matches the counts less closely (Pearson $0.677$, Spearman $0.659$): it separates rare facts from popular ones but orders facts within a tier unreliably. The two proxies agree with each other at Spearman $0.596$ and assign the same rare/popular label to $84\%$ of facts, which is the only property AdaPop uses.

\begin{table}[ht]
\centering
\footnotesize
\setlength{\tabcolsep}{3pt}
\begin{tabularx}{\linewidth}{|>{\raggedright\arraybackslash}X|c|c|}
\hline
\textbf{Proxy} & \textbf{Pearson} & \textbf{Spearman} \\ \hline
Wikidata score & 0.970 & 0.839 \\ \hline
\rowcolor{gray!10}
LLM judge & 0.677 & 0.659 \\ \hline
\end{tabularx}
\caption{Agreement of each popularity proxy with measured corpus frequency on DUET, counted over Pile-train ($383$B tokens) with infini-gram. Pearson is computed in log space. The two proxies agree with each other at Spearman $0.596$ and assign the same rare/popular label to $84\%$ of facts.}
\label{tab:proxy_corr}
\end{table}

\paragraph{Label-noise stress test.}
Correlations bound how far the proxies differ from each other; they do not show what happens when the signal is actively wrong. We therefore corrupt the Wikidata score directly and re-run AdaPop on DUET with Llama-3.1-8B-Instruct: first with multiplicative log-normal noise ($\sigma = 1.0$), then by randomly swapping the rare and popular labels of $50\%$ and of $100\%$ of facts. Results are in Table~\ref{tab:noise_stress}.

\begin{table}[ht]
\centering
\footnotesize
\setlength{\tabcolsep}{3pt}
\begin{tabularx}{\linewidth}{|>{\raggedright\arraybackslash}X|c|c|c|}
\hline
\textbf{Score noise} & \textbf{F. rare}~$\downarrow$ & \textbf{F. pop.}~$\downarrow$ & \textbf{R.}~$\uparrow$ \\ \hline
none & 0.05 & 0.04 & 0.96 \\ \hline
\rowcolor{gray!10}
log-normal, $\sigma{=}1.0$ & 0.02 & 0.05 & 0.96 \\ \hline
labels 50\% swapped & 0.02 & 0.05 & 0.97 \\ \hline
\rowcolor{gray!10}
labels 100\% swapped & 0.01 & 0.05 & 0.92 \\ \hline
\end{tabularx}
\caption{AdaPop under corrupted popularity scores (DUET, Llama-3.1-8B-Instruct, $\text{lr}=10^{-4}$). Forget ROUGE-L is reported per popularity tier. Retain degrades by $0.04$ even when every label is inverted.}
\label{tab:noise_stress}
\end{table}

Forget and retain remain stable under realistic noise, and a full label inversion degrades gracefully rather than breaking: retain holds at $0.92$, because the dual-ascent controller raises $\alpha$ when it detects fast retain degradation, pulling the run back toward the non-adaptive baseline. Two properties of the design account for this. Disagreements between proxies concentrate where $\beta_i \approx 1$, where the weighting is near-uniform in any case, and clipping $\beta_i$ to $[0.05, 2.0]$ bounds the influence of any single mis-scored outlier. In practice, checking two proxies against each other flags an unreliable signal before training rather than after.

\paragraph{Annotation and training cost.}
Popularity annotation is one-time preprocessing: $\approx$12 minutes for $2{,}000$ facts via the Wikidata sitelinks API, with $(a,b)$ available in closed form; the LLM-judge alternative is described in Appendix~\ref{app:llm_judge}. Both scale linearly in forget-set size. Per training step, AdaPop matches WGA, because $w_{i,t}$ reuses forward-pass probabilities under stop-gradient and adds no extra pass; the controller adds one retain evaluation and a two-scalar update per epoch. On a matched setup (Llama/DUET, one A100), training takes $29.0$ minutes for AdaPop against $29.4$ (WGA), $27.1$ (GD), $23.4$ (RMU), and $40.0$ (NPO). Since the controller also removes the per-dataset search over $\alpha$, total compute to a deployed checkpoint is lower than for a hand-tuned baseline.

\section{Qualitative Examples of Unlearning Generations}
\label{app:generation_examples}

We present qualitative examples of text generated by each algorithm at $\text{lr}=10^{-4}$. Outputs exhibiting token repetition or linguistic collapse are marked with \loopmark.

\vspace{1em}

\begin{genbox}[Example 1]
\textbf{Input:} \texttt{What is located in the administrative territorial entity of Mpwapwa?} \\
\textbf{Ground Truth:} \texttt{Mpwapwa District}

\tcblower

\textbf{Gemma-7B-it}
\begin{itemize}
    \item \textbf{GA:} \texttt{dises dises} \loopmark
    \item \textbf{GD:} \texttt{solidar emphat emphat} \loopmark
    \item \textbf{NPO:} \texttt{Tanzania}
    \item \textbf{WGA:} \texttt{Tanzania}
    \item \textbf{AdaPop:} \texttt{Tanzania Dodoma Region}
\end{itemize}

\textbf{Llama-3.1-8B-Instruct}
\begin{itemize}
    \item \textbf{GA:} \texttt{,,,,,,,} \loopmark
    \item \textbf{GD:} \texttt{,,,,,,,} \loopmark
    \item \textbf{NPO:} \texttt{Tanzania}
    \item \textbf{WGA:} \texttt{Tanzania}
    \item \textbf{AdaPop:} \texttt{Tanzania}
\end{itemize}

\textbf{Qwen2.5-7B-Instruct}
\begin{itemize}
    \item \textbf{GA:} \texttt{is is is} \loopmark
    \item \textbf{GD:} \texttt{is is is} \loopmark
    \item \textbf{NPO:} \texttt{Dodoma Region}
    \item \textbf{WGA:} \texttt{Dodoma Region}
    \item \textbf{AdaPop:} \texttt{Dodoma Region}
\end{itemize}
\end{genbox}

\vspace{1.5em}

\begin{genbox}[Example 2]
\textbf{Input:} \texttt{What is the capital of Sanniki?} \\
\textbf{Ground Truth:} \texttt{Gmina Sanniki}

\tcblower

\textbf{Gemma-7B-it}
\begin{itemize}
    \item \textbf{GA:} \texttt{dises dises} \loopmark
    \item \textbf{GD:} \texttt{emphat emphat} \loopmark
    \item \textbf{NPO:} \texttt{Lublin Voivodeship}
    \item \textbf{WGA:} \texttt{Lower vhs vhs vhs} \loopmark
    \item \textbf{AdaPop:} \texttt{Lublin}
\end{itemize}

\textbf{Llama-3.1-8B-Instruct}
\begin{itemize}
    \item \textbf{GA:} \texttt{,,,,,,,} \loopmark
    \item \textbf{GD:} \texttt{,,,,,,,} \loopmark
    \item \textbf{NPO:} \texttt{Gmina Sanniki, Żuromin County, Masovian Voivodeship, northeastern Poland}
    \item \textbf{WGA:} \texttt{Maz Pi Pi Pi} \loopmark
    \item \textbf{AdaPop:} \texttt{Lodski}
\end{itemize}

\textbf{Qwen2.5-7B-Instruct}
\begin{itemize}
    \item \textbf{GA:} \texttt{is is is} \loopmark
    \item \textbf{GD:} \texttt{is is is} \loopmark
    \item \textbf{NPO:} \texttt{Gmina Sanniki is a town in Poland and is located in the Lubusz Voivodeship. It is not a city with a capital.}
    \item \textbf{WGA:} \texttt{Lublin Voievodeship}
    \item \textbf{AdaPop:} \texttt{Georgia is divided into 9 regions, Sannake is not among them, there is no city called Sanniki as capital city is Tbilisi.}
\end{itemize}
\end{genbox}

\section{Comparison with Additional Unlearning Methods}
\label{app:additional_baselines}

We evaluate eleven methods not included in the main comparison: UNDIAL~\cite{dong2025undial}, RMU~\cite{li2024wmdp}, PDU~\cite{entesariconstrained}, NPO-SAM~\cite{fan2025towards}, SimNPO~\cite{fan2026simplicity}, SatImp~\cite{yangexploring}, Adaptive RMU~\cite{dang2025effects}, AltPO~\cite{mekala2025alternate}, FLAT~\cite{wang2025llm}, TPO~\cite{zhou2026not}, and CE-U~\cite{yang2025u}. All use identical LoRA configuration ($r{=}32$, $\alpha_{\mathrm{LoRA}}{=}64$, lr $= 10^{-4}$, 5 epochs). Tables~\ref{tab:baselines-llama}--\ref{tab:baselines-gemma} report ROUGE-L Recall; $^\dag$marks model collapse (F\,$\approx$\,0, R\,$\approx$\,0).

Three patterns emerge. First, distributional methods (UNDIAL, RMU) produce forget ROUGE-L of $0.734$--$0.884$ on DUET, barely below the model before unlearning, because their self-distillation and representation-misdirection signals do not address parametric encoding depth of entity facts; on RWKU they achieve partial erasure but remain well above \textbf{AdaPop} (e.g., Llama: UNDIAL $0.568$, RMU $0.801$ vs.\ $0.078$). Second, PDU's dual controller reduces forget to $0.476$--$0.677$ on DUET but incurs a measurable retain penalty (down to $0.710$ on Qwen/RWKU), showing that constrained optimisation alone is insufficient without popularity-calibrated gradient scaling. Third, preference-based variants (SimNPO, SatImp, AltPO, TPO) preserve retention ($\geq 0.891$) but under-erase popular facts (DUET forget ROUGE-L $0.070$--$0.899$, against $0.023$--$0.056$ for \textbf{AdaPop}), inheriting NPO's slow convergence on deeply memorised content. NPO-SAM erases only partially (forget $0.211$--$0.684$) and loses retention while doing so (down to $0.596$ on Qwen/RWKU). Adaptive RMU partially recovers on RWKU but at a large retain cost (retain $0.466$--$0.819$), and fails on DUET ($0.293$--$0.933$). FLAT and CE-U collapse across all settings ($^\dag$).

\textbf{AdaPop} achieves the lowest forget ROUGE-L among the stable methods in Tables~\ref{tab:baselines-llama}--\ref{tab:baselines-gemma} on both benchmarks (e.g., Llama/DUET: $0.043$; Qwen/RWKU: $0.016$) while maintaining retain ROUGE-L $\geq 0.855$ on every model-benchmark combination.

\begin{table}[ht]
\centering
\footnotesize
\setlength{\tabcolsep}{3pt}
\begin{tabularx}{\linewidth}{|l|>{\centering\arraybackslash}X|c|c|}
\hline
\multirow{2}{*}{\textbf{Bench.}} & \multirow{2}{*}{\textbf{Algorithm}} & \multicolumn{2}{c|}{\textbf{ROUGE-L}} \\ \cline{3-4}
 & & \textbf{F.} $\downarrow$ & \textbf{R.} $\uparrow$ \\ \hline
\rowcolor{cyan!15} \cellcolor{white} & \textbf{w/o MU}    & 0.939 & 0.968 \\ \cline{2-4}
\rowcolor{violet!15} \cellcolor{white} & \textbf{AdaPop} & \underline{0.043} & 0.959 \\ \cline{2-4}
\rowcolor{gray!10} \cellcolor{white} & UNDIAL           & 0.884 & 0.998 \\ \cline{2-4}
\cellcolor{white} & RMU                              & 0.882 & 0.998 \\ \cline{2-4}
\rowcolor{gray!10} \cellcolor{white} & PDU              & 0.476 & 0.954 \\ \cline{2-4}
\cellcolor{white} & NPO-SAM                          & 0.684 & 0.970 \\ \cline{2-4}
\rowcolor{gray!10} \cellcolor{white} & SimNPO           & 0.340 & 0.999 \\ \cline{2-4}
\cellcolor{white} & SatImp                           & 0.108 & 0.995 \\ \cline{2-4}
\rowcolor{gray!10} \cellcolor{white} & Adaptive RMU     & 0.933 & 0.963 \\ \cline{2-4}
\cellcolor{white} & AltPO                            & 0.181 & \underline{1.000} \\ \cline{2-4}
\rowcolor{gray!10} \cellcolor{white} & FLAT$^\dag$      & 0.001 & 0.001 \\ \cline{2-4}
\cellcolor{white} & TPO                              & 0.397 & 0.997 \\ \cline{2-4}
\multirow{-13}{*}{\textbf{DUET}} \cellcolor{white} & CE-U$^\dag$                      & 0.000 & 0.000 \\ \hline
\rowcolor{cyan!15} \cellcolor{white} & \textbf{w/o MU}    & 0.755 & 0.827 \\ \cline{2-4}
\rowcolor{violet!15} \cellcolor{white} & \textbf{AdaPop} & \underline{0.078} & 0.972 \\ \cline{2-4}
\rowcolor{gray!10} \cellcolor{white} & UNDIAL           & 0.568 & 0.939 \\ \cline{2-4}
\cellcolor{white} & RMU                              & 0.801 & 0.986 \\ \cline{2-4}
\rowcolor{gray!10} \cellcolor{white} & PDU              & 0.201 & 0.909 \\ \cline{2-4}
\cellcolor{white} & NPO-SAM                          & 0.529 & 0.885 \\ \cline{2-4}
\rowcolor{gray!10} \cellcolor{white} & SimNPO           & 0.573 & 0.987 \\ \cline{2-4}
\cellcolor{white} & SatImp                           & 0.417 & 0.985 \\ \cline{2-4}
\rowcolor{gray!10} \cellcolor{white} & Adaptive RMU     & 0.162 & 0.819 \\ \cline{2-4}
\cellcolor{white} & AltPO                            & 0.200 & \underline{0.988} \\ \cline{2-4}
\rowcolor{gray!10} \cellcolor{white} & FLAT$^\dag$      & 0.004 & 0.003 \\ \cline{2-4}
\cellcolor{white} & TPO                              & 0.341 & 0.986 \\ \cline{2-4}
\multirow{-13}{*}{\textbf{RWKU}} \cellcolor{white} & CE-U$^\dag$                      & 0.000 & 0.000 \\ \hline
\end{tabularx}
\caption{ROUGE-L Recall for additional baselines, Llama-3.1-8B-Instruct at $\text{lr}=10^{-4}$. \textbf{w/o MU} = the original checkpoint, with no unlearning applied; \textbf{AdaPop} reproduced from Table~\ref{tab:rouge-llama} for reference. F.~$\downarrow$ = forget; R.~$\uparrow$ = retain. \underline{Underline} = best among non-collapsed methods. $^\dag$Model collapse.}
\label{tab:baselines-llama}
\end{table}

\begin{table}[ht]
\centering
\footnotesize
\setlength{\tabcolsep}{3pt}
\begin{tabularx}{\linewidth}{|l|>{\centering\arraybackslash}X|c|c|}
\hline
\multirow{2}{*}{\textbf{Bench.}} & \multirow{2}{*}{\textbf{Algorithm}} & \multicolumn{2}{c|}{\textbf{ROUGE-L}} \\ \cline{3-4}
 & & \textbf{F.} $\downarrow$ & \textbf{R.} $\uparrow$ \\ \hline
\rowcolor{cyan!15} \cellcolor{white} & \textbf{w/o MU}    & 0.921 & 0.886 \\ \cline{2-4}
\rowcolor{violet!15} \cellcolor{white} & \textbf{AdaPop} & \underline{0.056} & 0.950 \\ \cline{2-4}
\rowcolor{gray!10} \cellcolor{white} & UNDIAL           & 0.797 & 0.930 \\ \cline{2-4}
\cellcolor{white} & RMU                              & 0.734 & 0.983 \\ \cline{2-4}
\rowcolor{gray!10} \cellcolor{white} & PDU              & 0.632 & 0.869 \\ \cline{2-4}
\cellcolor{white} & NPO-SAM                          & 0.211 & 0.769 \\ \cline{2-4}
\rowcolor{gray!10} \cellcolor{white} & SimNPO           & 0.295 & 0.991 \\ \cline{2-4}
\cellcolor{white} & SatImp                           & 0.070 & 0.985 \\ \cline{2-4}
\rowcolor{gray!10} \cellcolor{white} & Adaptive RMU     & 0.562 & 0.831 \\ \cline{2-4}
\cellcolor{white} & AltPO                            & 0.247 & \underline{0.999} \\ \cline{2-4}
\rowcolor{gray!10} \cellcolor{white} & FLAT$^\dag$      & 0.000 & 0.001 \\ \cline{2-4}
\cellcolor{white} & TPO                              & 0.460 & 0.999 \\ \cline{2-4}
\multirow{-13}{*}{\textbf{DUET}} \cellcolor{white} & CE-U$^\dag$                      & 0.001 & 0.001 \\ \hline
\rowcolor{cyan!15} \cellcolor{white} & \textbf{w/o MU}    & 0.570 & 0.666 \\ \cline{2-4}
\rowcolor{violet!15} \cellcolor{white} & \textbf{AdaPop} & \underline{0.016} & 0.855 \\ \cline{2-4}
\rowcolor{gray!10} \cellcolor{white} & UNDIAL           & 0.313 & 0.668 \\ \cline{2-4}
\cellcolor{white} & RMU                              & 0.407 & 0.922 \\ \cline{2-4}
\rowcolor{gray!10} \cellcolor{white} & PDU              & 0.455 & 0.710 \\ \cline{2-4}
\cellcolor{white} & NPO-SAM                          & 0.298 & 0.596 \\ \cline{2-4}
\rowcolor{gray!10} \cellcolor{white} & SimNPO           & 0.338 & 0.934 \\ \cline{2-4}
\cellcolor{white} & SatImp                           & 0.187 & 0.891 \\ \cline{2-4}
\rowcolor{gray!10} \cellcolor{white} & Adaptive RMU     & 0.118 & 0.466 \\ \cline{2-4}
\cellcolor{white} & AltPO                            & 0.170 & \underline{0.989} \\ \cline{2-4}
\rowcolor{gray!10} \cellcolor{white} & FLAT$^\dag$      & 0.004 & 0.003 \\ \cline{2-4}
\cellcolor{white} & TPO                              & 0.382 & 0.934 \\ \cline{2-4}
\multirow{-13}{*}{\textbf{RWKU}} \cellcolor{white} & CE-U$^\dag$                      & 0.003 & 0.001 \\ \hline
\end{tabularx}
\caption{ROUGE-L Recall for additional baselines, Qwen2.5-7B-Instruct at $\text{lr}=10^{-4}$. Notation as in Table~\ref{tab:baselines-llama}.}
\label{tab:baselines-qwen}
\end{table}

\begin{table}[ht!]
\centering
\footnotesize
\setlength{\tabcolsep}{3pt}
\begin{tabularx}{\linewidth}{|l|>{\centering\arraybackslash}X|c|c|}
\hline
\multirow{2}{*}{\textbf{Bench.}} & \multirow{2}{*}{\textbf{Algorithm}} & \multicolumn{2}{c|}{\textbf{ROUGE-L}} \\ \cline{3-4}
 & & \textbf{F.} $\downarrow$ & \textbf{R.} $\uparrow$ \\ \hline
\rowcolor{cyan!15} \cellcolor{white} & \textbf{w/o MU}    & 0.892 & 0.926 \\ \cline{2-4}
\rowcolor{violet!15} \cellcolor{white} & \textbf{AdaPop} & \underline{0.023} & 0.976 \\ \cline{2-4}
\rowcolor{gray!10} \cellcolor{white} & UNDIAL           & 0.819 & 0.980 \\ \cline{2-4}
\cellcolor{white} & RMU                              & 0.851 & 0.992 \\ \cline{2-4}
\rowcolor{gray!10} \cellcolor{white} & PDU              & 0.677 & 0.934 \\ \cline{2-4}
\cellcolor{white} & NPO-SAM                          & 0.539 & 0.777 \\ \cline{2-4}
\rowcolor{gray!10} \cellcolor{white} & SimNPO           & 0.581 & 0.997 \\ \cline{2-4}
\cellcolor{white} & SatImp                           & 0.070 & 0.998 \\ \cline{2-4}
\rowcolor{gray!10} \cellcolor{white} & Adaptive RMU     & 0.293 & 0.828 \\ \cline{2-4}
\cellcolor{white} & AltPO                            & 0.262 & \underline{1.000} \\ \cline{2-4}
\rowcolor{gray!10} \cellcolor{white} & FLAT$^\dag$      & 0.001 & 0.001 \\ \cline{2-4}
\cellcolor{white} & TPO                              & 0.899 & 0.999 \\ \cline{2-4}
\multirow{-13}{*}{\textbf{DUET}} \cellcolor{white} & CE-U$^\dag$                      & 0.000 & 0.000 \\ \hline
\rowcolor{cyan!15} \cellcolor{white} & \textbf{w/o MU}    & 0.471 & 0.551 \\ \cline{2-4}
\rowcolor{violet!15} \cellcolor{white} & \textbf{AdaPop} & \underline{0.034} & 0.948 \\ \cline{2-4}
\rowcolor{gray!10} \cellcolor{white} & UNDIAL           & 0.453 & 0.830 \\ \cline{2-4}
\cellcolor{white} & RMU                              & 0.547 & 0.968 \\ \cline{2-4}
\rowcolor{gray!10} \cellcolor{white} & PDU              & 0.404 & 0.825 \\ \cline{2-4}
\cellcolor{white} & NPO-SAM                          & 0.271 & 0.644 \\ \cline{2-4}
\rowcolor{gray!10} \cellcolor{white} & SimNPO           & 0.373 & 0.969 \\ \cline{2-4}
\cellcolor{white} & SatImp                           & 0.102 & 0.964 \\ \cline{2-4}
\rowcolor{gray!10} \cellcolor{white} & Adaptive RMU     & 0.091 & 0.535 \\ \cline{2-4}
\cellcolor{white} & AltPO                            & 0.172 & \underline{0.989} \\ \cline{2-4}
\rowcolor{gray!10} \cellcolor{white} & FLAT$^\dag$      & 0.005 & 0.004 \\ \cline{2-4}
\cellcolor{white} & TPO                              & 0.594 & 0.971 \\ \cline{2-4}
\multirow{-13}{*}{\textbf{RWKU}} \cellcolor{white} & CE-U$^\dag$                      & 0.000 & 0.000 \\ \hline
\end{tabularx}
\caption{ROUGE-L Recall for additional baselines, Gemma-7B-it at $\text{lr}=10^{-4}$. Notation as in Table~\ref{tab:baselines-llama}.}
\label{tab:baselines-gemma}
\end{table}

\section{General Capability Preservation: Extended Discussion}
\label{app:lm_eval}

We measure capability preservation with MMLU (factual recall across 57 domains) and HellaSwag (sentence completion requiring sustained coherence). Table~\ref{tab:lm_eval} (main paper, §\ref{sec:lm_eval}) reports results for all methods at $\text{lr}=10^{-4}$, averaged across DUET and RWKU checkpoints.

\paragraph{GA collapses; GD shows asymmetric damage.}
GA destroys both benchmarks across all three models (MMLU $0.23$--$0.25$ vs.\ $0.47$--$0.71$ before unlearning). GD preserves MMLU (within $0.01$ of the starting value on Llama/Qwen) but degrades HellaSwag (Llama $0.73 \to 0.59$; Gemma $0.64 \to 0.37$): independent factual recall is reinforced by the retain objective, but sustained-context reasoning is more sensitive to insufficient retain regularisation.

\paragraph{NPO, WGA, AdaPop preserve capabilities.}
All three remain within $0.05$ MMLU of the pre-unlearning value across architectures. WGA and \textbf{AdaPop} perform supervised descent on the retain split (continual-learning--style replay), which can leave the model slightly better calibrated on held-out items: WGA matches or exceeds the starting HellaSwag score, and \textbf{AdaPop} does the same on Llama and Qwen (small drop on Gemma, $0.62$ vs.\ $0.64$). NPO stays at or slightly below it. Retain-split hidden-state cosine for \textbf{AdaPop} is comparable to WGA and NPO (Table~\ref{tab:internal_agg}), confirming output calibration rather than representational drift.

These results indicate that targeted factual unlearning and general capability preservation are orthogonal concerns when retain regularisation is applied consistently. The primary risk is over-aggressive forgetting (GA, GD), not popularity-aware gradient calibration.

\section{Evaluation Metric Definitions}
\label{app:metric_formulas}

This appendix gives the exact form of the metrics summarised in §\ref{sec:experiments}. Let $\theta_0$ denote the parameters before unlearning and $\theta$ those after it. For an evaluation example $i$ with prompt $x_i$ and gold answer $y_i$, let $A_i$ be the set of answer-token positions ($|A_i|$ tokens; prompt tokens are masked out), and write $q_\theta(y_{i,t}) = q_\theta(y_{i,t} \mid x_i, y_{i,<t})$. We write $\mathcal{V}$ for the vocabulary and $L$ for the number of transformer layers. All four internal metrics are computed per example and then averaged over the split, so each reported value is a mean over examples.

\paragraph{Output-level metrics.}
ROUGE-L Recall~\cite{lin2004rouge} is the longest-common-subsequence recall of the generated answer $\hat{y}_i$ against the gold answer $y_i$:
\begin{equation}
\mathrm{ROUGE\text{-}L}_i = \frac{\left|\mathrm{LCS}(\hat{y}_i,\, y_i)\right|}{|y_i|}.
\end{equation}
Cosine Similarity embeds both strings with a sentence encoder $\phi$~\cite{reimers2019sentence} and compares them:
\begin{equation}
\mathrm{CosSim}_i = \frac{\phi(\hat{y}_i)^\top \phi(y_i)}{\lVert \phi(\hat{y}_i) \rVert \, \lVert \phi(y_i) \rVert}.
\end{equation}

\paragraph{$\Delta$LP.}
The gold answer's log-probability is summed over its tokens, and the metric is the shift in that quantity:
\begin{equation}
\Delta\mathrm{LP}_i = \sum_{t \in A_i} \log q_{\theta}(y_{i,t}) \;-\; \sum_{t \in A_i} \log q_{\theta_0}(y_{i,t}).
\end{equation}
Because the sum runs over the whole answer, $\Delta$LP scales with answer length; it is comparable across methods on a fixed split, not across splits.

\paragraph{$\Delta$Rank.}
The rank of a gold token is one plus the number of vocabulary items the model scores strictly above it,
\begin{equation}
r_\theta(i,t) = 1 + \left| \left\{ v \in \mathcal{V} : q_\theta(v \mid x_i, y_{i,<t}) > q_\theta(y_{i,t}) \right\} \right|,
\end{equation}
so $r_\theta = 1$ means the gold token is the argmax. Ranks are averaged over the answer tokens of an example, and $\Delta$Rank is the shift in that average:
\begin{equation}
\Delta\mathrm{Rank}_i = \frac{1}{|A_i|}\sum_{t \in A_i} \Big( r_{\theta}(i,t) - r_{\theta_0}(i,t) \Big).
\end{equation}
Values are large in absolute terms because vocabularies contain $10^5$ or more tokens.

\paragraph{Hid.Cos.}
Let $h^{(l)}_\theta(i,t)$ be the hidden state at layer $l$ and position $t$. We average the last four layers, mean-pool over answer tokens,
\begin{equation}
\bar{h}_\theta(i) = \frac{1}{|A_i|}\sum_{t \in A_i}\; \frac{1}{4}\sum_{l = L-3}^{L} h^{(l)}_\theta(i,t),
\end{equation}
and compare the pooled vectors before and after unlearning:
\begin{equation}
\mathrm{Hid.Cos}_i = \frac{\bar{h}_{\theta_0}(i)^\top \bar{h}_{\theta}(i)}{\lVert \bar{h}_{\theta_0}(i) \rVert \, \lVert \bar{h}_{\theta}(i) \rVert}.
\end{equation}

\paragraph{KL.}
The divergence is taken from the pre-unlearning distribution to the unlearned one, at each answer position over the full vocabulary, and averaged over the answer:
\begin{equation}
\mathrm{KL}_i = \frac{1}{|A_i|}\sum_{t \in A_i} \sum_{v \in \mathcal{V}} q_{\theta_0}(v) \log \frac{q_{\theta_0}(v)}{q_{\theta}(v)},
\end{equation}
where $q_\cdot(v)$ abbreviates $q_\cdot(v \mid x_i, y_{i,<t})$. Unlike $\Delta$LP, which tracks only the gold continuation, KL registers any redistribution of probability mass at those positions.

\section{Internal Representation Metrics}
\label{app:internal_metrics}

\begin{figure}[h!]
    \centering
    \includegraphics[width=\columnwidth]{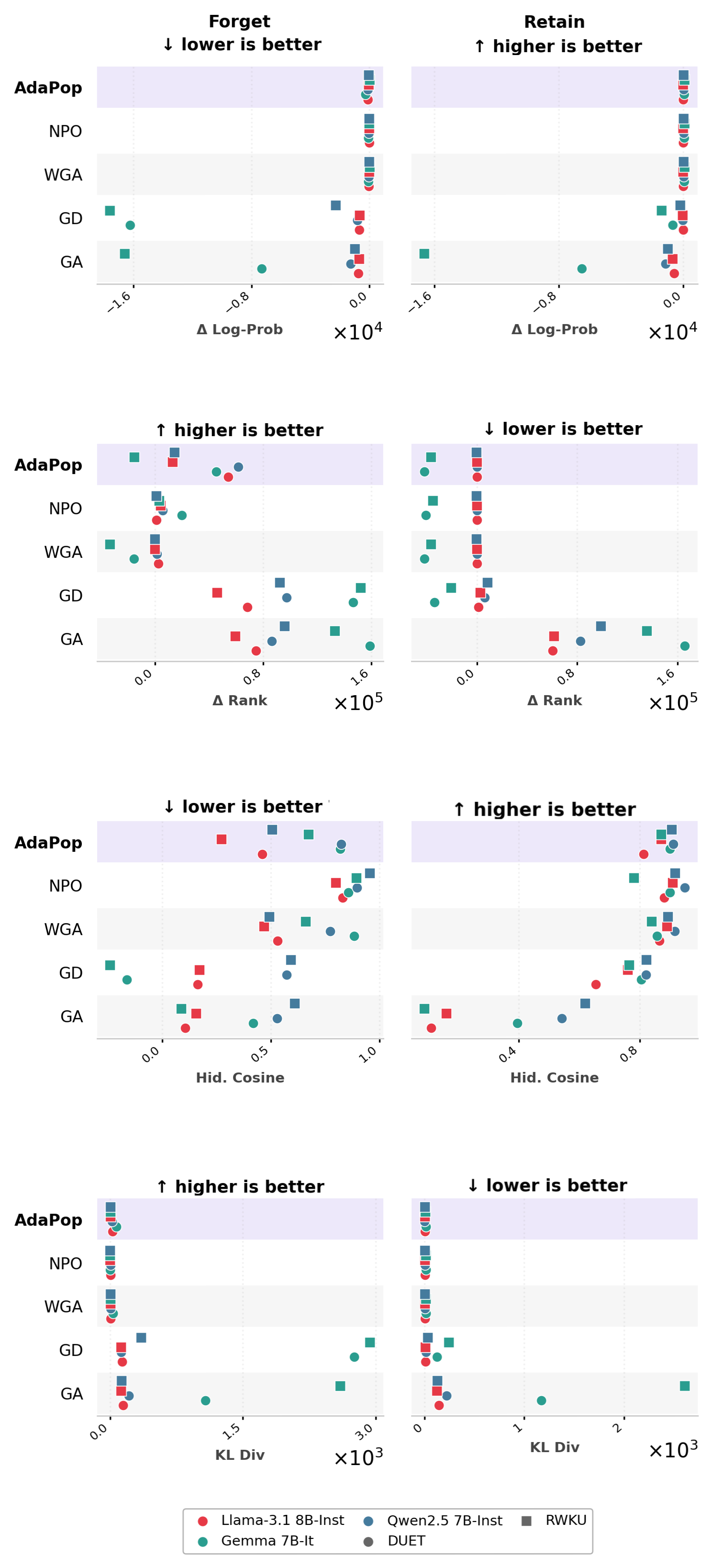}
    \caption{Internal representation metrics at $\text{lr}=10^{-4}$, across three models. Each row corresponds to one metric; each column corresponds to one data split. For the forget split, deeper erasure corresponds to lower $\Delta$LP, higher $\Delta$Rank, lower Hid.Cos, and higher KL; directional desiderata are reversed for the retain split. Marker shape indicates benchmark (DUET vs.\ RWKU); colour indicates model family.}
    \label{fig:internal_metrics}
\end{figure}

Figure~\ref{fig:internal_metrics} shows the per-model breakdown of all four internal metrics ($\Delta$LP, $\Delta$Rank, Hid.Cos, KL) at $\text{lr}=10^{-4}$, complementing the model-averaged results in Table~\ref{tab:internal_agg}.

\section{Model-Specific Observations}
\label{app:model_specific}

\paragraph{Llama-3.1-8B-Instruct.}
Cleanest forget-retain separation. \textbf{AdaPop} reduces forget ROUGE-L from $0.939$ to $0.043$ on DUET (retain $0.959$) and from $0.755$ to $0.078$ on RWKU (retain $0.972$, the strongest RWKU retain across architectures). WGA is marginally lower on DUET paraphrase ($0.042$ vs.\ $0.045$); AdaPop separates on RWKU adversarial robustness ($0.262$ vs.\ $0.396$) and $\Delta$Rank depth (Table~\ref{tab:internal_agg}).

\paragraph{Gemma-7B-it.}
Gemma has the smallest pre-unlearning DUET cosine ($0.586$), reflecting distinctive token-probability calibration. \textbf{AdaPop} drops forget cosine to $0.206$ (DUET) and $0.068$ (RWKU) while retain cosine rises to $0.976$ and $0.961$, above the retain values before unlearning; this reflects retain-loss minimisation during training rather than representational change (Appendix~\ref{app:lm_eval}). On robustness: DUET paraphrase $0.027$ vs.\ WGA $0.050$; RWKU adversarial $0.195$ vs.\ $0.239$.

\paragraph{Qwen2.5-7B-Instruct.}
Largest \textbf{AdaPop} gain on RWKU: forget $0.570 \to 0.016$, retain $0.666 \to 0.855$, retain cosine $0.347 \to 0.911$. On DUET, AdaPop and WGA are similar at the surface ($0.056$ vs.\ $0.058$ forget) but separate under robustness: DUET paraphrase $0.074$ vs.\ $0.104$; RWKU adversarial $0.144$ vs.\ $0.211$, indicating WGA's Qwen forgetting is surface-level.

\section{Design Motivation: Gradient-Based Unlearning and the Retain Controller}
\label{app:design_motivation}

\paragraph{Why the ascent family.}
Gradient-based unlearning methods split into two families. The \emph{ascent family} (GA, GD, WGA) directly maximises forget-set NLL; the \emph{preference family} (NPO and variants) reframes unlearning as alignment, replacing the unbounded ascent signal with a bounded preference-margin objective. The two families exhibit opposite failure modes: ascent risks retain-set damage if not regularised, while preference-based methods converge slowly on deeply memorised facts because the KL constraint to the reference model bounds the per-step update magnitude. We adopt the ascent family because, empirically, the primary failure mode in factual unlearning is \emph{insufficient forgetting}: a method that preserves retain but leaves popular facts recoverable under paraphrase or adversarial prompts has not unlearned in any practical sense. Our experiments (§\ref{sec:results}) confirm this: NPO maintains near-perfect retain across all settings but under-erases popular DUET facts (forget ROUGE-L $0.626$--$0.684$, recoverable under adversarial queries).

Ascent is the right family because its weakness is separable from its strength. The retain-side damage can be handled by a regulariser \emph{decoupled} from the forget objective, which is the role of the dual-ascent controller. That decoupling lets AdaPop treat forget aggressiveness (via $\beta_i$) and retain stability (via $\alpha$) as two dedicated components instead of compromising both in a single objective.

The robustness evaluation (Table~\ref{tab:robustness}) provides direct empirical support. WGA's DUET paraphrase ROUGE-L exceeds AdaPop's on Qwen ($0.104$ vs.\ $0.074$) and Gemma ($0.050$ vs.\ $0.027$), and its adversarial-attack ROUGE-L exceeds AdaPop's across all three models ($0.211$--$0.396$ vs.\ $0.144$--$0.262$). The gap is largest on adversarial reformulations, which probe parametric encoding rather than surface output, consistent with the view that confidence-based weighting suppresses tokens the model currently outputs but leaves the underlying encoding largely intact, whereas popularity-based weighting targets the encoding itself.

\subsection{Dynamic Retain Coefficient as Constrained Optimisation}
\label{app:design_motivation:dual}

GD and WGA use a fixed retain coefficient $\alpha$ that must be tuned per (dataset, model) pair: too small and retain quality degrades, too large and well-memorised popular facts survive erasure~\cite{bu2025unlearning}. The principled formulation is to treat the retain term as an inequality constraint rather than a fixed penalty:
\begin{equation}
    \min_{\theta} \; L_f(\theta) \quad \text{s.t.} \quad L_r(\theta) - L_r(\theta_0) \;\le\; \varepsilon,
\label{eq:constrained}
\end{equation}
where $\theta_0$ is the pre-unlearning reference checkpoint and $L_r$ is the retain loss. The Lagrangian is
\begin{align}
    \mathcal{L}(\theta, \lambda) &\;=\; L_f(\theta) \nonumber\\
    &\quad+\; \lambda \left(L_r(\theta) - L_r(\theta_0) - \varepsilon\right), \qquad \lambda \ge 0.
\end{align}
\citet{entesariconstrained} adopt the same constrained formulation for LLM unlearning. AdaPop's implementation differs in two practical respects:
\begin{enumerate}
    \item \textbf{Epoch-level dual updates.} $\lambda$ is updated once per epoch (not per step), reducing oscillation from noisy per-batch retain losses. The drift signal $\delta_k$ is computed as a relative deviation from the epoch-1 retain loss $R_{\mathrm{ref}} = R_1$ rather than from the initial checkpoint, to absorb the unavoidable first-epoch shift from LoRA fine-tuning:
    \begin{gather}
        \delta_k = \max\!\left(0,\;\frac{R_k - R_{\mathrm{ref}}}{\max(R_{\mathrm{ref}}, \xi)}\right),\\
        \lambda_{k+1} = \Pi_{[0,\lambda_{\max}]}\!\left(\lambda_k + \eta_\lambda(\delta_k - \varepsilon)\right).
    \end{gather}
    \item \textbf{Popularity-weighted forget objective.} The controller is applied to the popularity-weighted forget loss $L_f$ (Eq.~\ref{eq:forget_loss}), rather than unweighted NLL ascent. This composition lets the two components address distinct empirical failure modes, verified in Appendix~\ref{app:ablation}.
\end{enumerate}

\paragraph{Stability under non-convexity.}
Standard primal-dual ascent has well-understood convergence behaviour only on convex programs. The LLM training objective is non-convex, so we do not claim formal convergence; the controller is designed for empirical stability instead. Three design choices mitigate the typical risks (limit cycles, divergence under aggressive $\eta_\lambda$): (i) $\lambda$ updates at epoch rather than batch granularity, so $\delta_k$ averages over many gradient steps; (ii) $\lambda$ is clipped to $[0, \lambda_{\max}]$ with $\lambda_{\max} = 5.0$; (iii) $\lambda_0 = 0$, so the effective retain weight is $\alpha_0 = 0.5$ at initialisation, matching a standard WGA-style retain penalty. Across the full $3{\times}2$ (model, benchmark) grid and the learning-rate sweep in Appendix~\ref{app:lr_sensitivity}, the controller keeps $|\delta_k| \le \varepsilon = 0.1$ at every epoch, and retain-split $\Delta$LP and Hid.Cos for \textbf{AdaPop} match WGA (Table~\ref{tab:internal_agg}).

\paragraph{Empirical decoupling of the two components.}
The popularity exponent and the retain controller act on distinct aspects of the gradient: $\beta$ reshapes the per-token weighting \emph{within} the forget set (via \eqref{eq:effective-weight}), while $\alpha$ scales the retain term \emph{relative} to the forget term in the combined update $\theta_{k+1} - \theta_k = -\eta_\theta \,(\nabla L_f^{\beta} + \alpha \nabla L_r)$. They are not strictly orthogonal in the linear-algebra sense ($\beta$ affects the overall forget-gradient scale as well as its direction), but they address \emph{empirically distinct failure modes}: removing the controller breaks retain stability under increasing forget pressure (a magnitude problem), while flattening $\beta$ to a constant breaks erasure of popular facts (a per-token weighting problem). The $2{\times}2$ ablation in Appendix~\ref{app:ablation} verifies this decomposition: fixing either component leaves the corresponding failure mode unaddressed, and only the full configuration achieves both deep erasure and stable retention across the tested learning-rate range.

\section{Component Ablation: Dual-Ascent Controller vs.\ Popularity Exponent}
\label{app:ablation}

The design rationale of AdaPop (Appendix~\ref{app:design_motivation:dual}) attributes two distinct empirical failure modes to two distinct components: the retain controller addresses magnitude failure (the forget gradient overwhelming a fixed retain penalty), while the popularity exponent addresses per-token weighting failure (uniform pressure on all tokens regardless of memorisation depth). We test this decomposition with a $2{\times}2$ ablation that independently varies whether $\alpha$ is fixed at $0.5$ or dual-ascent--controlled, and whether $\beta_i$ is fixed at $0.1$ (the same exponent across all facts, near-GA regime) or set by the popularity power-law:
\begin{itemize}
    \item $\alpha{=}0.5,\,\beta{=}0.1$: fixed retain coefficient, fixed exponent (uniform across facts).
    \item $\alpha{=}0.5,\,\beta{=}\text{dyn}$: fixed retain coefficient, popularity-based $\beta_i$ (isolates the exponent without the controller).
    \item $\alpha{=}\text{dyn},\,\beta{=}0.1$: dual-ascent controller, fixed exponent (isolates the controller without popularity weighting).
    \item $\alpha{=}\text{dyn},\,\beta{=}\text{dyn}$: full AdaPop.
\end{itemize}

\begin{figure}[t]
    \centering
    \includegraphics[width=\linewidth]{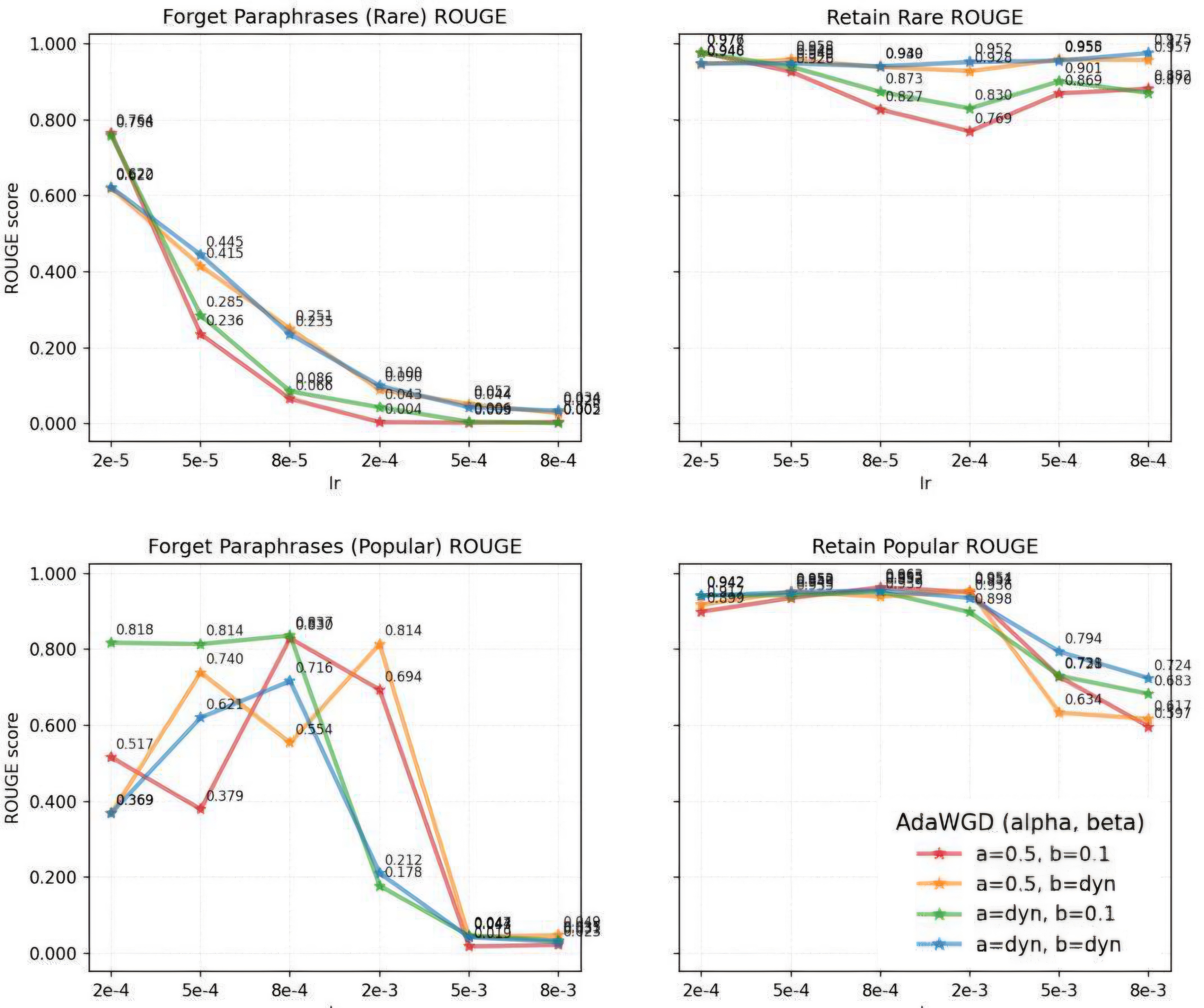}
    \caption{Component ablation on DUET. \textit{Top row}: rare-fact paraphrase forget ROUGE-L ($\downarrow$) and retain ROUGE-L ($\uparrow$) over $\text{lr} \in [2\times10^{-5}, 8\times10^{-4}]$. \textit{Bottom row}: popular-fact paraphrase forget ROUGE-L ($\downarrow$) and retain ROUGE-L ($\uparrow$) over $\text{lr} \in [2\times10^{-4}, 8\times10^{-3}]$.}
    \label{fig:ablation}
\end{figure}

The decomposition predicts that fixing either component leaves its failure mode unaddressed: variants without dyn-$\alpha$ should lose retain quality under increasing forget pressure regardless of $\beta$, and variants without dyn-$\beta$ should under-erase popular facts regardless of $\alpha$. Figure~\ref{fig:ablation} reports paraphrase forget ROUGE-L ($\downarrow$) and retain ROUGE-L ($\uparrow$) by popularity tier across learning rates on DUET. The rare-fact plots span $\text{lr} \in [2\times10^{-5},\,8\times10^{-4}]$; popular-fact plots span $[2\times10^{-4},\,8\times10^{-3}]$. The order-of-magnitude gap in the learning rate required for comparable forgetting between tiers is itself direct evidence of the popularity gap: popular facts are substantially more resistant to gradient-based erasure.

\paragraph{Controller drives retain stability.}
Without the controller ($\alpha{=}0.5$), retain on rare facts drops to $0.769$ at $\text{lr}{=}2\times10^{-4}$, while both dyn-$\alpha$ variants maintain $\geq 0.927$. On popular facts at higher rates, fixed-$\alpha$ variants fall to $\leq 0.634$ at $\text{lr}{=}5\times10^{-3}$, while at that rate the controller keeps retain above $0.72$. The forget gradient eventually overwhelms any fixed retain penalty, regardless of how it is distributed across tokens by $\beta$. The popularity exponent alone does not address this failure: it only redistributes gradient mass \emph{within} the forget set, leaving the forget-vs-retain magnitude ratio uncontrolled.

\paragraph{Popularity exponent improves the Pareto frontier.}
At the learning rates where both dyn-$\alpha$ variants achieve comparable forget ROUGE-L ($\approx 0.02$ on popular facts), full AdaPop ($\alpha{=}\text{dyn},\,\beta{=}\text{dyn}$) retains $0.724$ versus $0.683$ for the controller-only variant ($\alpha{=}\text{dyn},\,\beta{=}0.1$). The mechanism follows from Proposition~\ref{prop:regimes}: $\beta_i < 1$ for popular facts allocates per-token effective weight $p_{i,t}^{\beta_i - 1}$ proportionally to remaining confidence, concentrating updates on tokens that still resist erasure rather than ones that have already lost confidence. This makes forgetting more targeted, allowing the controller to apply less overall retain pressure for the same erasure depth.

\paragraph{Fixed $\alpha$ with dynamic $\beta$ has no viable operating point.}
$\alpha{=}0.5,\,\beta{=}\text{dyn}$ achieves the highest popular-paraphrase forget ROUGE-L across the entire learning-rate sweep, reaching $0.814$ at $\text{lr}{=}2\times10^{-3}$, close to NPO's under-erased popular tier ($0.854$ on Llama, Table~\ref{tab:tier_paraphrase}). For popular facts, $\beta_i \ll 1$ already produces a softly distributed per-token signal; without the controller adapting $\alpha$ as forget pressure increases, the retain penalty constrains the update before deep erasure is reached. This confirms that the popularity exponent and the dual-ascent controller are complementary: the exponent shapes how gradient mass is distributed across tokens, but the controller is required to let the forget loss apply enough overall pressure to overcome popular-fact memorisation.

\paragraph{Summary.}
The ablation shows that each component is necessary: the controller alone (dyn-$\alpha$, $\beta=0.1$) achieves retain stability but cannot reach deep forgetting of popular facts; the exponent alone (fixed-$\alpha$, $\beta=\text{dyn}$) has no viable operating point, losing retain at the learning rates that erase popular facts and under-erasing at the rates where retain holds. Only the full $(\alpha{=}\text{dyn},\,\beta{=}\text{dyn})$ configuration achieves low forget and high retain simultaneously across the tested learning-rate range.

\section{Learning-Rate Sensitivity and Algorithm Stability}
\label{app:lr_sensitivity}

\begin{figure}[t]
    \centering
    \includegraphics[width=\linewidth]{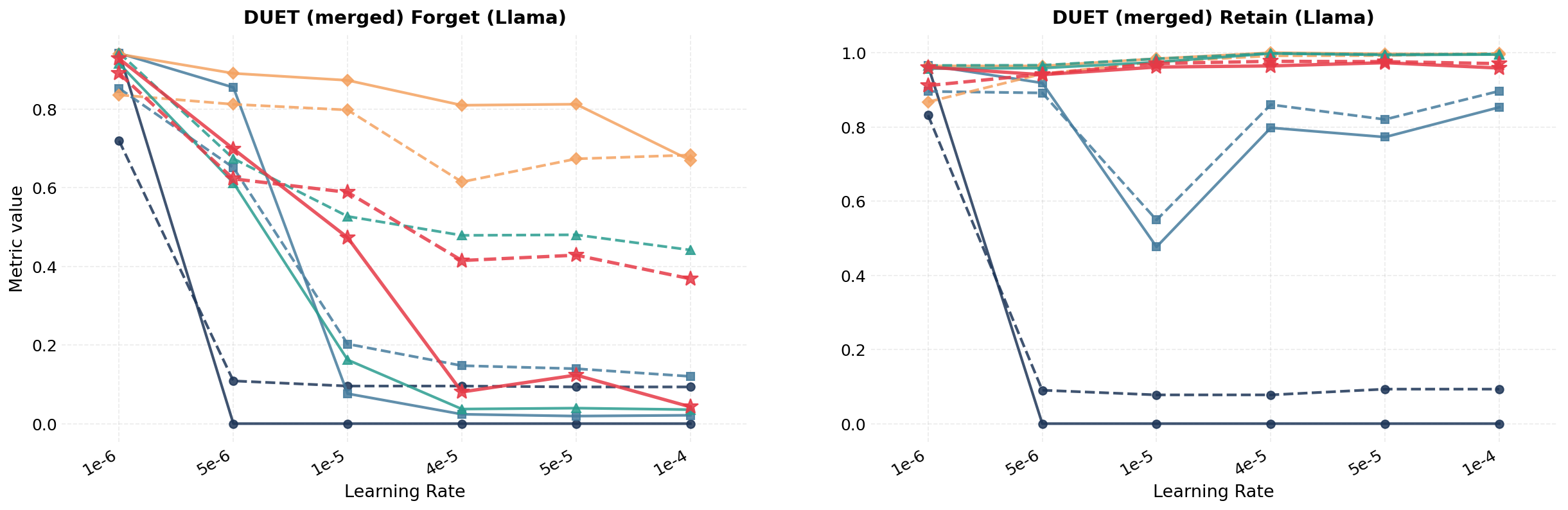}
    \includegraphics[width=\linewidth]{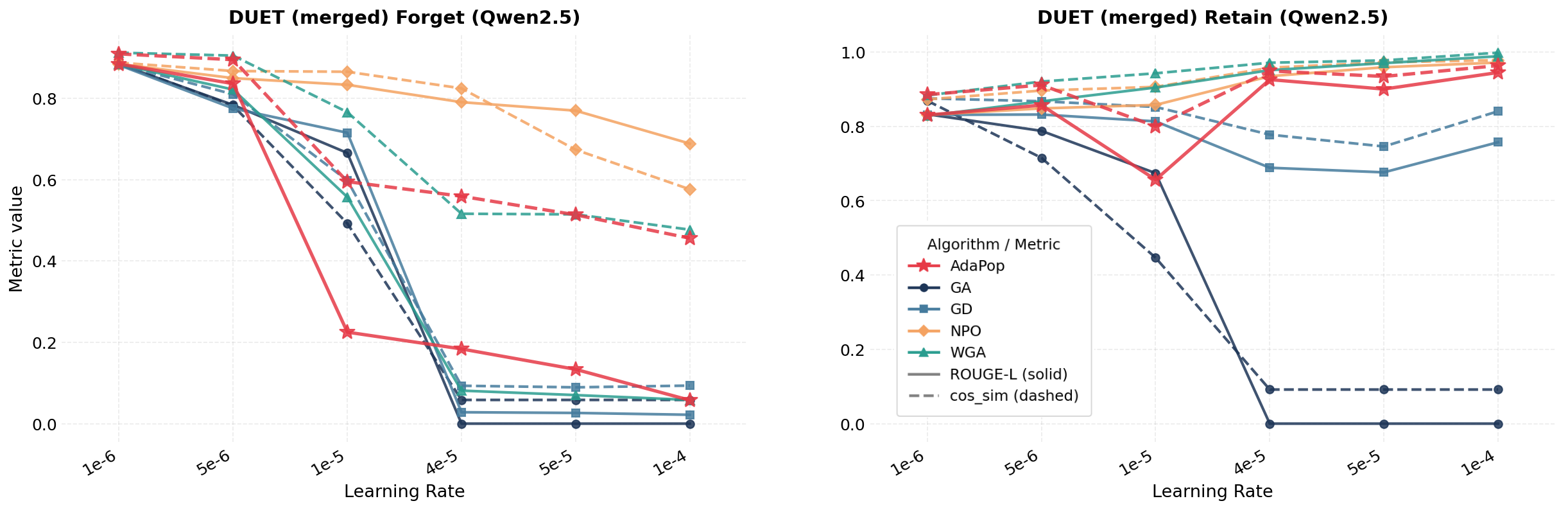}
    \includegraphics[width=\linewidth]{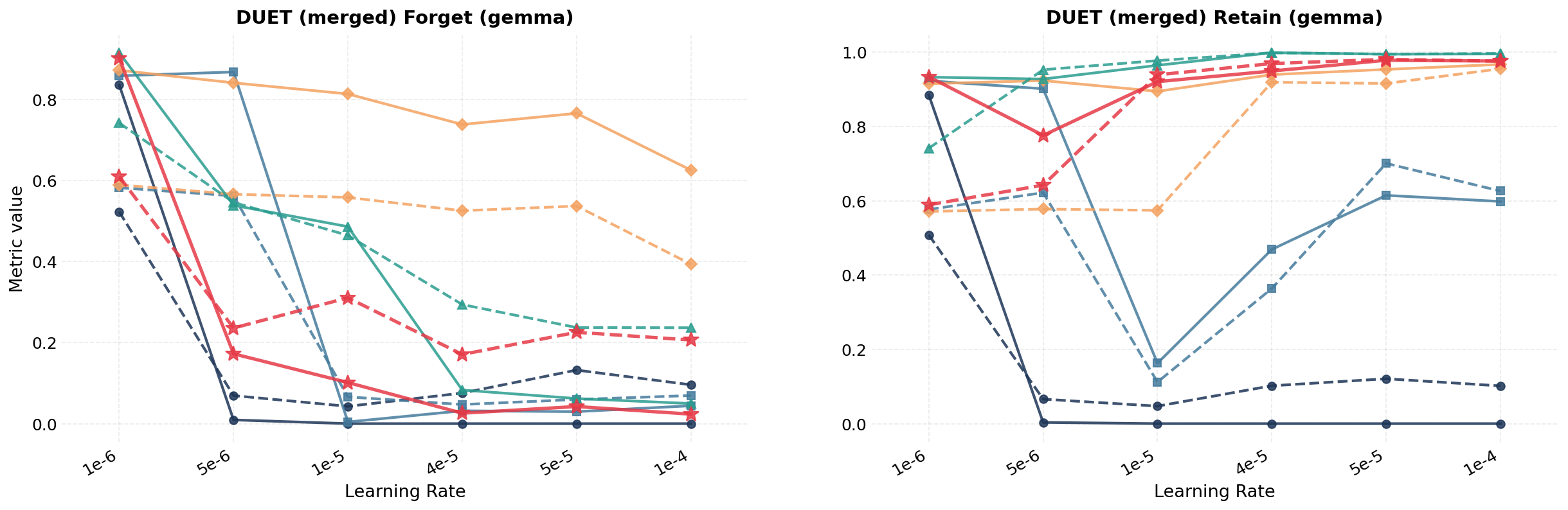}
    \caption{ROUGE-L (solid) and Cosine Similarity (dashed) on the merged DUET forget (left) and retain (right) splits across learning rates, for Llama (top), Qwen (middle), and Gemma (bottom).}
    \label{fig:lr_duet}
\end{figure}

\begin{figure}[t]
    \centering
    \includegraphics[width=\linewidth]{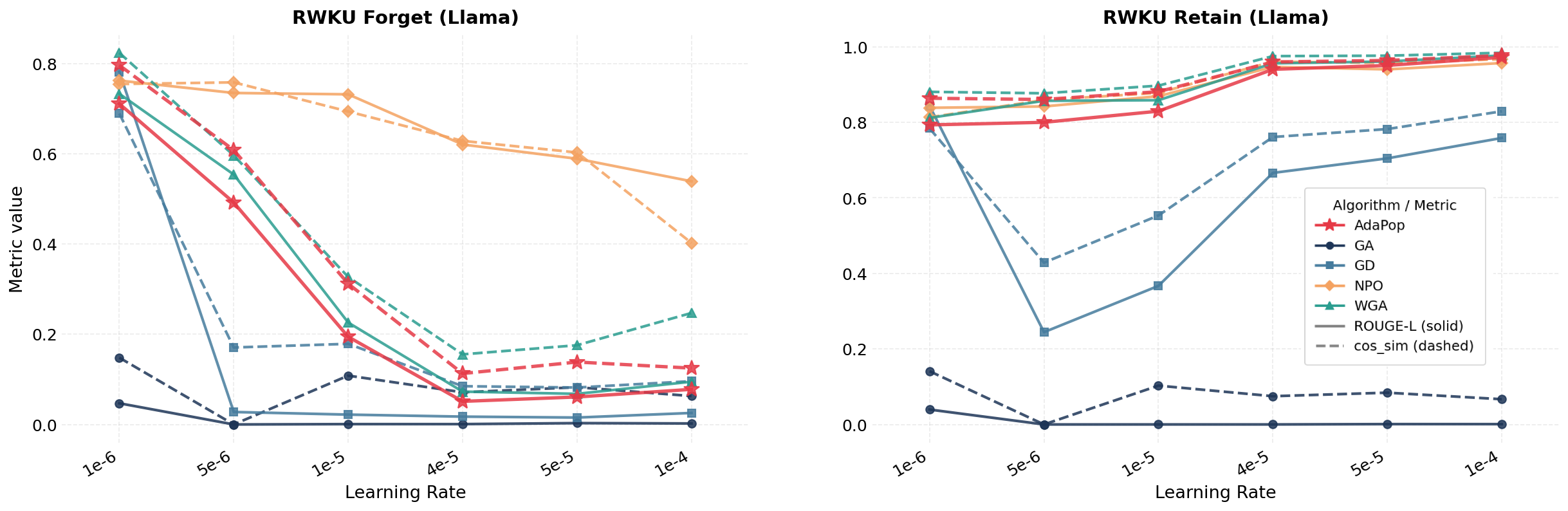}
    \includegraphics[width=\linewidth]{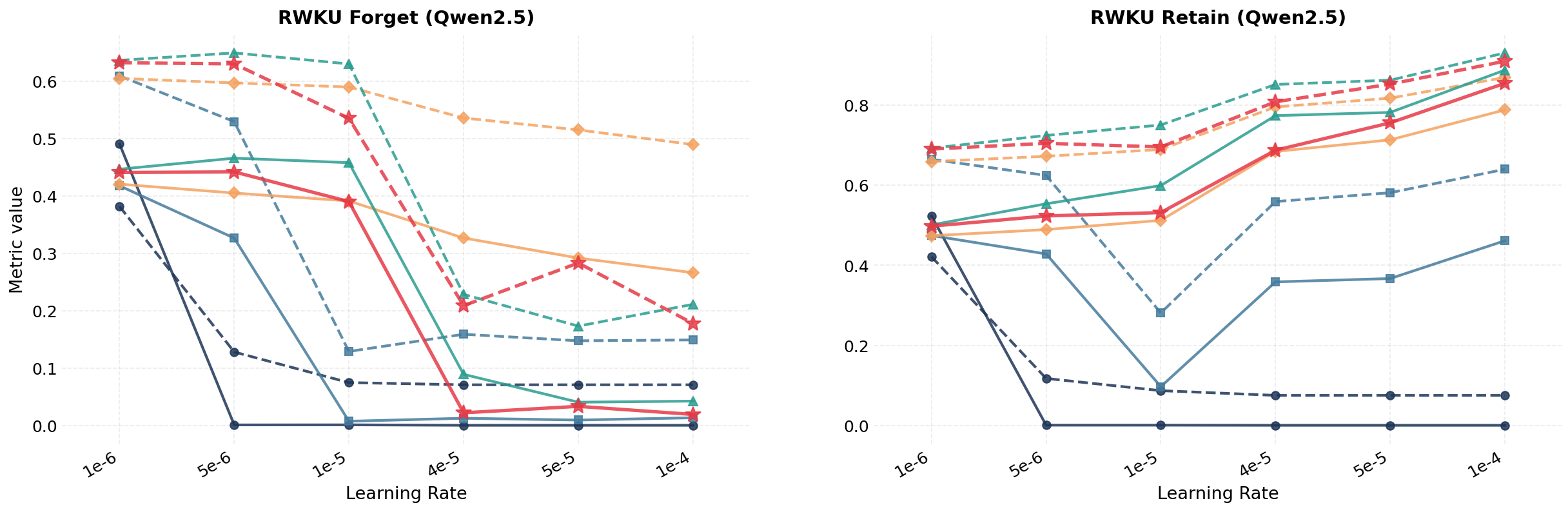}
    \includegraphics[width=\linewidth]{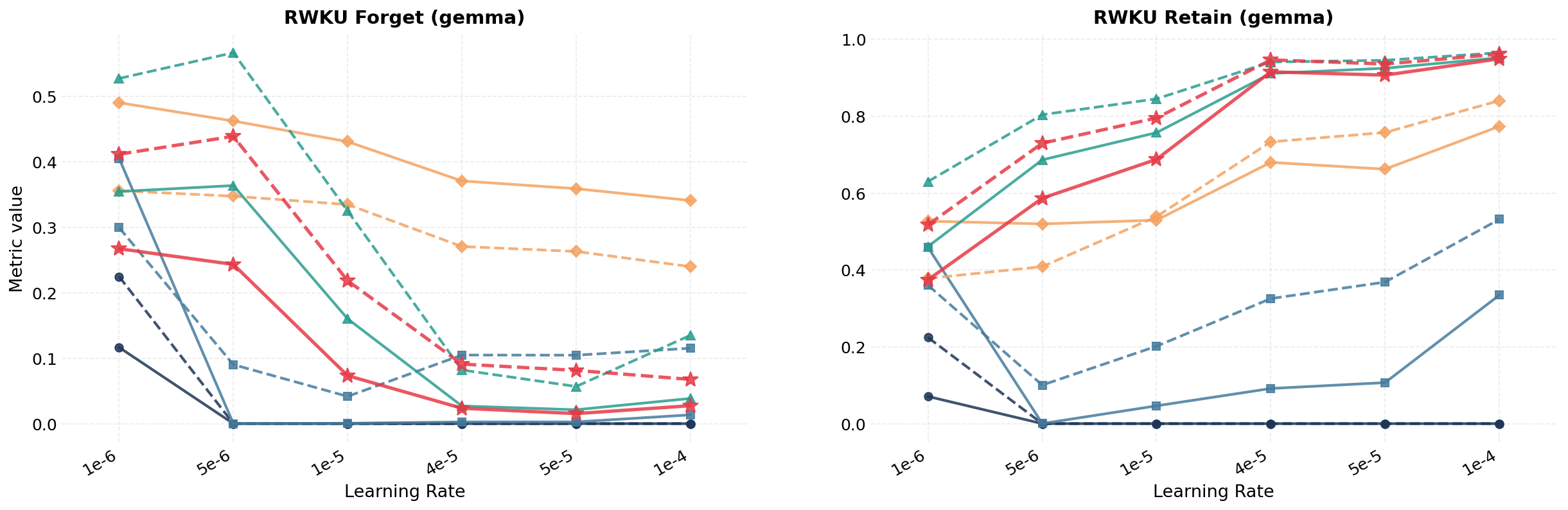}
    \caption{ROUGE-L (solid) and Cosine Similarity (dashed) on the RWKU forget (left) and retain (right) splits across learning rates, for Llama (top), Qwen (middle), and Gemma (bottom).}
    \label{fig:lr_rwku}
\end{figure}

\begin{figure}[t]
    \centering
    \includegraphics[width=\linewidth]{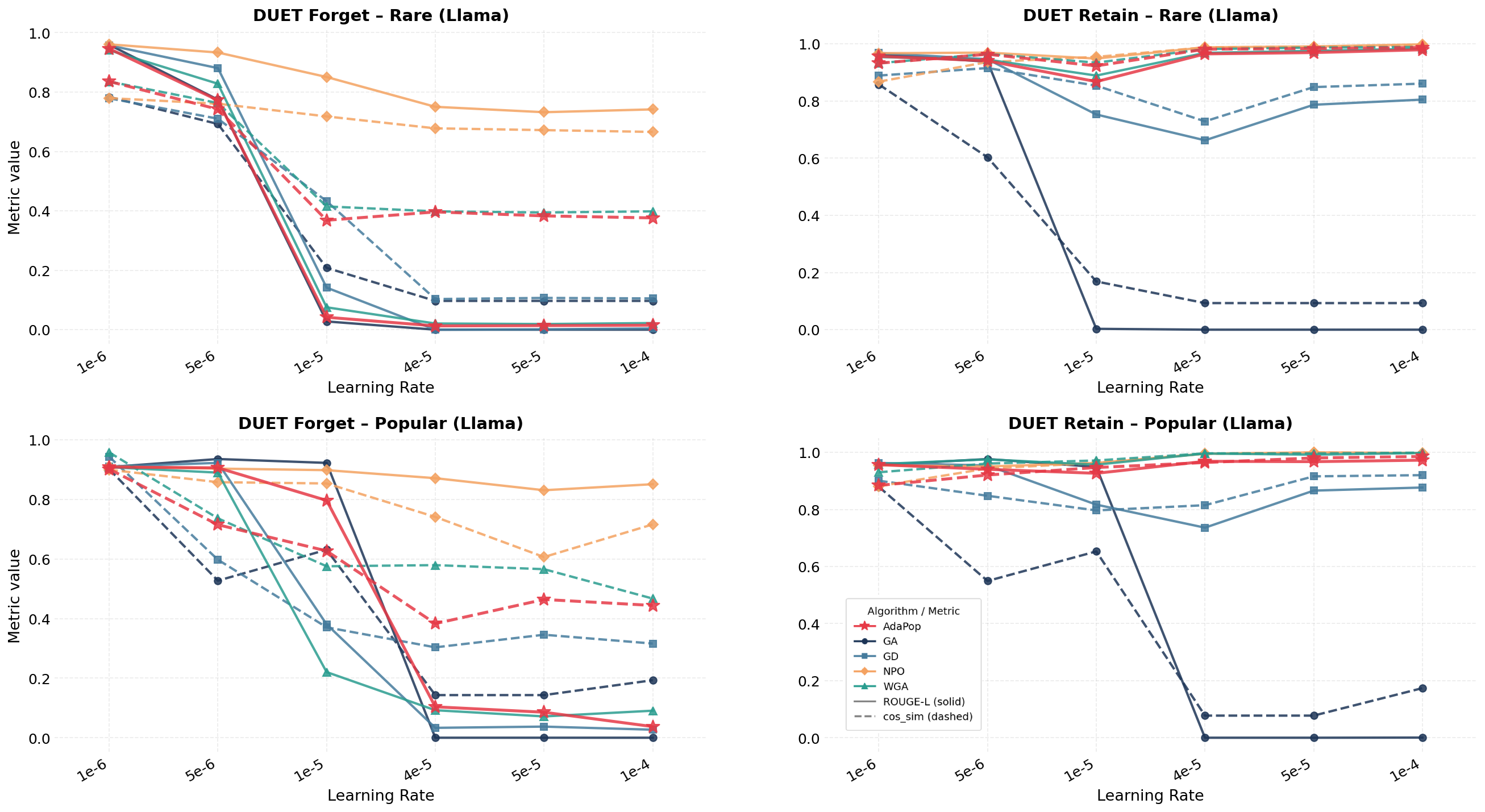}
    \caption{DUET forget and retain splits separated by popularity tier (Llama). \textit{Top row}: rare facts. \textit{Bottom row}: popular facts. The order-of-magnitude difference in the learning rate required to achieve comparable forgetting across the two tiers is direct evidence of the popularity gap.}
    \label{fig:lr_duet_split}
\end{figure}

We characterise each method's forget-retain trade-off across $\text{lr} \in \{10^{-6}, 5{\times}10^{-6}, 10^{-5}, 5{\times}10^{-5}, 10^{-4}\}$ to identify stable operating regions. Figures~\ref{fig:lr_duet} and~\ref{fig:lr_rwku} plot ROUGE-L (solid) and Cosine Similarity (dashed) on the merged forget and retain splits; Figure~\ref{fig:lr_duet_split} separates DUET into rare and popular tiers for Llama.

\paragraph{GA, GD: no viable operating region.}
GA drives forget and retain ROUGE-L to zero at every tested rate on DUET. GD is marginally better at $10^{-6}$ but loses retain at any rate that achieves meaningful forgetting. Both are unusable above $10^{-6}$ on RWKU as well.

\paragraph{NPO: stable retain, weak forget on DUET.}
NPO retain is near-perfect across all rates, but DUET forget curves are nearly flat with learning rate, showing the preference signal cannot overcome high parametric memorisation. On RWKU (lower baseline memorisation), NPO achieves reasonable forgetting at higher rates, but forget cosine remains the highest among non-collapsing methods, indicating shallow representational change.

\paragraph{WGA: model-dependent stability.}
WGA is competitive at $\text{lr}=10^{-4}$ but unstable at higher rates on Qwen and Gemma (retain degrades visibly at $\text{lr} \geq 5{\times}10^{-5}$); forget cosine does not decrease proportionally to ROUGE-L, indicating surface suppression. Without an automatic retain controller, WGA's optimal rate must be grid-searched per (model, benchmark).

\paragraph{AdaPop: monotonic forget with controlled retain.}
AdaPop is the only method whose forget quality improves monotonically with learning rate on both benchmarks and all models. The controller raises the retain coefficient to compensate for the stronger forget signal; on RWKU, retain ROUGE-L actually rises with learning rate (retain-loss minimisation, Appendix~\ref{app:lm_eval}). Forget cosine decreases in parallel with ROUGE-L, confirming representational rather than surface-level erasure.

\paragraph{Popularity-tier asymmetry on DUET (Figure~\ref{fig:lr_duet_split}).}
For Llama, rare facts reach near-zero forget at $\text{lr} \approx 5{\times}10^{-5}$ but popular facts require $\text{lr} \approx 10^{-4}$; even there, WGA's popular-fact forget cosine stays above AdaPop's, and NPO's popular-tier forget curves remain flat at the level before unlearning. The popularity gap is intrinsic to fact encoding, not an artefact of any single learning-rate choice.

\paragraph{Stability to popularity distribution and batch composition.}
\textbf{AdaPop} is robust to both the dataset-level popularity distribution and per-batch composition. DUET spans nearly two orders of magnitude in popularity score ($69$--$3{,}763$) while RWKU is narrower ($0$--$704$); WGA's confidence-based weighting under-shoots the forget signal on RWKU's lower-memorisation facts, whereas \textbf{AdaPop}'s per-fact exponent adjusts automatically. The merged batches (Figure~\ref{fig:lr_duet}) and tier-pure batches (Figure~\ref{fig:lr_duet_split}) produce consistent curves: the epoch-level controller observes retain-loss drift and adjusts $\alpha$ regardless of which popularity tier dominates. Under fixed $\alpha$ (Appendix~\ref{app:ablation}), homogeneous popular-fact batches cause retain collapse at moderate learning rates.

\section{Per-Epoch Unlearning Dynamics}
\label{app:epoch_dynamics}

\begin{figure}[H]
    \centering
    \includegraphics[width=\linewidth]{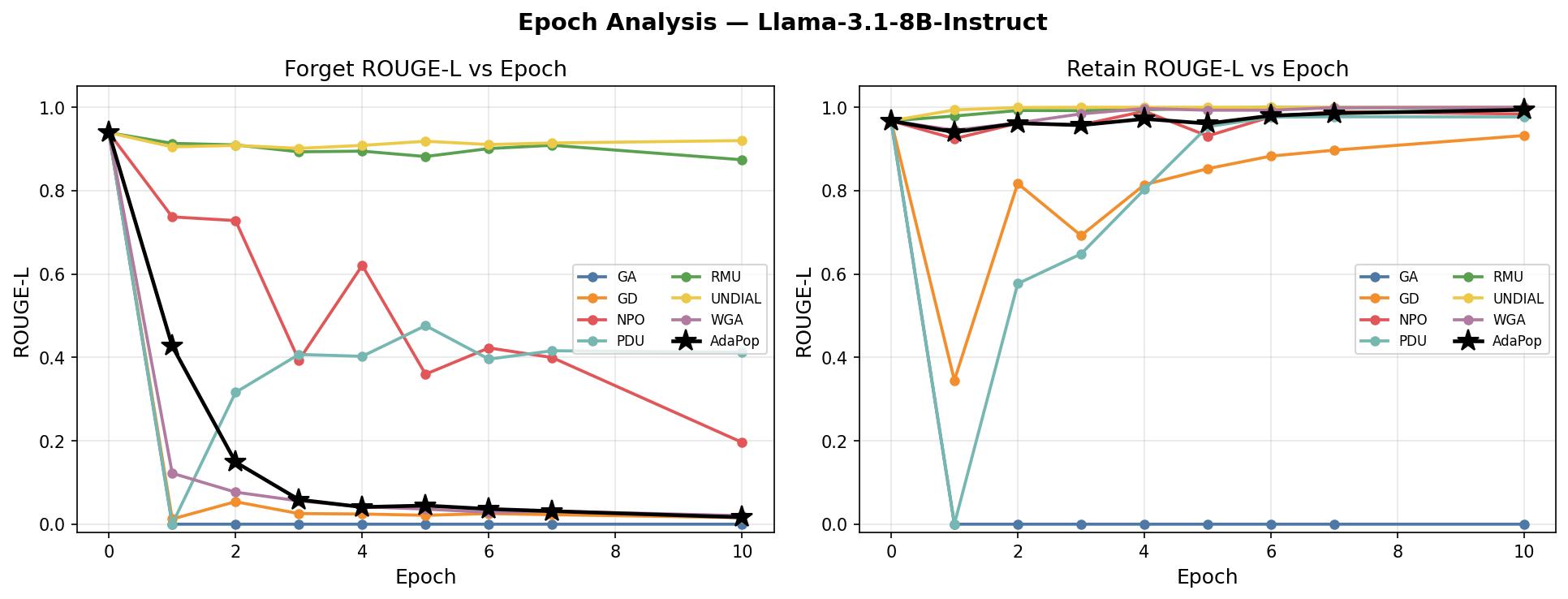}
    \caption{Forget (left) and retain (right) ROUGE-L Recall per training epoch on DUET, Llama-3.1-8B-Instruct, at $\text{lr}=10^{-4}$.}
    \label{fig:epoch_dynamics}
\end{figure}

Figure~\ref{fig:epoch_dynamics} traces forget and retain ROUGE-L Recall across training epochs (0--10) on DUET for Llama-3.1-8B-Instruct at $\text{lr}=10^{-4}$, covering seven baseline algorithms. AdaPop is not included: its per-epoch behaviour is characterised separately through the learning-rate sweep (Appendix~\ref{app:lr_sensitivity}), where the dual-ascent dynamics are the primary object of study.

\paragraph{Collapse and transient instability (GA, GD, PDU).}
GA reaches forget $= 0.000$ and retain $= 0.000$ after epoch~1 and never recovers. GD achieves strong forgetting at epoch~1 (forget $0.012$) but retain crashes to $0.345$; subsequent epochs recover retain to $0.932$ while forget oscillates non-monotonically around its epoch-1 value, leaving practitioners reliant on careful early stopping. PDU collapses at epoch~1 (forget/retain $= 0.000/0.001$), recovers to $\approx 0.41/0.97$ by epoch~4, then stagnates with forget pinned above $0.39$.

\paragraph{Oscillation and shallow erasure (NPO, RMU, UNDIAL).}
NPO retain is stable ($\geq 0.924$) but forget oscillates between $0.196$ and $0.737$ across epochs 1--10 with no consistent downward trend, reflecting slow preference-alignment convergence on deeply memorised facts. RMU and UNDIAL never unlearn on entity-centric QA (forget $\geq 0.87$ at every epoch); both are designed for distributional/skill-based unlearning rather than fact-level erasure.

\paragraph{Monotonic convergence (WGA).}
WGA is the only baseline with smooth monotonic forgetting: forget drops from $0.122$ (epoch~1) to $0.020$ (epoch~10) on a single seed-42 trajectory while retain stays $\geq 0.945$. (The main-table value $0.036 \pm .002$ in Table~\ref{tab:rouge-llama} is the three-seed mean at the fixed 5-epoch stopping point used for all methods.) Any epoch beyond~4 is a viable operating point, motivating WGA as the primary comparison baseline.

Stable monotonic convergence is rare: GA, GD, and PDU exhibit collapse or transient instability; NPO oscillates; RMU and UNDIAL never erase entity-level facts. WGA is the only baseline that improves epoch-over-epoch without retain degradation. The dual-ascent controller in AdaPop is designed to close the stability gap that GD and NPO expose: treating the retain constraint as feedback prevents both early collapse and forget-curve oscillation.

\end{document}